\documentclass[11pt]{article}

\usepackage[preprint]{acl}

\usepackage{times}
\usepackage{latexsym}

\usepackage[T1]{fontenc}
\usepackage[utf8]{inputenc}
\usepackage{CJKutf8}  % Chinese character support
\usepackage{textcomp}
\usepackage{newunicodechar}
\newunicodechar{−}{\textminus}

\usepackage{amsmath}
\usepackage{amssymb}

\usepackage{graphicx}
\usepackage{booktabs}
\usepackage{multirow}
\usepackage{array}  % for m{} column type (vertical centering)
\usepackage{standalone}

\usepackage{enumitem}
\usepackage[table]{xcolor}  % table option enables \rowcolor
\usepackage{pifont}  % for \ding symbols

\usepackage{tikz}
\usetikzlibrary{shapes,shapes.geometric,arrows,arrows.meta,positioning,shadows,calc,fit,backgrounds,decorations.pathreplacing}
\usepackage{pgfplots}
\pgfplotsset{compat=1.18}
\PassOptionsToPackage{hyphens,spaces,obeyspaces}{url}

\usepackage{microtype}

\usepackage{placeins}
\usepackage{stfloats}  % Better control for figure* placement
\usepackage{algorithm}
\usepackage{algorithmic}

\newcommand{\frameworkurl}{https://github.com/yha9806/VULCA-Framework}

\newcommand{\dataseturl}{https://huggingface.co/datasets/harryHURRY/vulca-bench}
\title{Cross-Cultural Expert-Level Art Critique Evaluation with Vision-Language Models}

\author{
  Haorui Yu\textsuperscript{1},
  Xuehang Wen\textsuperscript{2},
  Fengrui Zhang\textsuperscript{3}, and
  Qiufeng Yi\textsuperscript{4}
\\
  \textsuperscript{1}DJCAD, University of Dundee, United Kingdom \\
  \textsuperscript{2}Hebei Academy of Fine Art, China \\
  \textsuperscript{3}Computer Science, Nanjing University, China \\
  \textsuperscript{4}Department of Mechanical Engineering, School of Engineering, University of Birmingham
}

\begin{document}

\maketitle

% =============================================================================
% Abstract
% =============================================================================
\begin{abstract}

Vision-Language Models (VLMs) excel at visual description yet remain under-validated for cultural interpretation. Existing benchmarks assess perception without interpretation, and common evaluation proxies, such as automated metrics and LLM-judge averaging, are unreliable for culturally sensitive generative tasks. We address this measurement gap with a tri-tier evaluation framework grounded in art-theoretical constructs (\S\ref{sec:related}). The framework operationalises cultural understanding through five levels (L1--L5) and 165 culture-specific dimensions across six traditions: Tier~I computes automated quality indicators, Tier~II applies rubric-based single-judge scoring, and Tier~III calibrates the aggregate score to human expert ratings via sigmoid calibration. Applied to 15 VLMs across 294 evaluation pairs, the validated instrument reveals that (i)~automated metrics and judge scoring measure different constructs, establishing single-judge calibration as the more reliable alternative; (ii)~cultural understanding degrades from visual description (L1--L2) to cultural interpretation (L3--L5); and (iii)~Western art samples consistently receive higher scores than non-Western ones. To our knowledge, this is the first cross-cultural evaluation instrument for generative art critique, providing a reproducible methodology for auditing VLM cultural competence.
Framework code is available at \url{\frameworkurl}.

\end{abstract}

% =============================================================================
% Main Content
% =============================================================================

% Section 1: Introduction
% =============================================================================
% Paper 1: Introduction Section (Compressed per D1-D3)
% =============================================================================

\section{Introduction}

% --- Move 1: Establishing Territory ---

\begin{figure}[t]
\centering
\includegraphics[width=0.85\columnwidth]{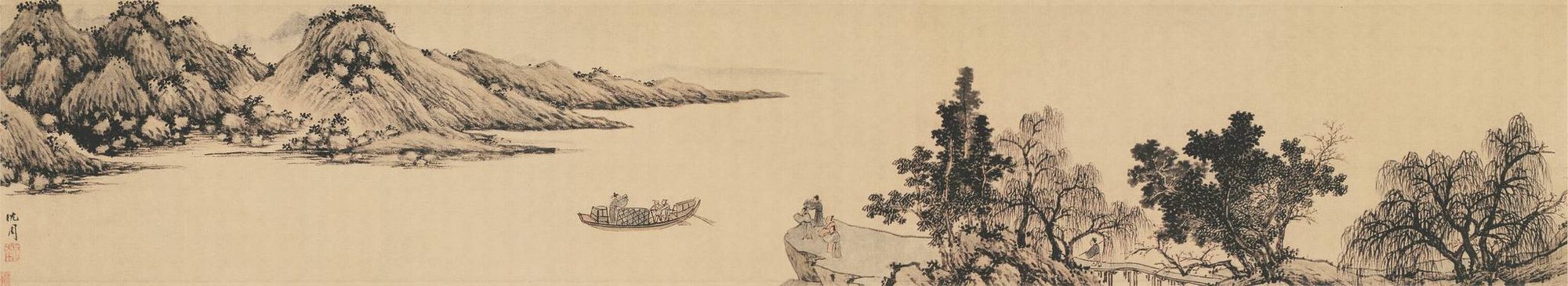}
\vspace{-0.5em}
\caption{VLM vs.\ Expert on Shen Zhou's \textit{Farewell at Jingjiang}. The VLM stops at generic L1--L2 description (score 2.45), while the expert spans L1--L5 with specific technique terms, historical context, and aesthetic interpretation.}
\label{fig:vlm_vs_expert}
\end{figure}

Vision-Language Models (VLMs) achieve high scores on perception benchmarks, yet these metrics mask a critical limitation: VLMs describe what they see but rarely interpret what it means. Art understanding requires moving from visual description (L1--L2) to cultural interpretation, including symbolism, historical context, and philosophical aesthetics (L3--L5, see \S\ref{sec:related} for theoretical grounding). Figure~\ref{fig:vlm_vs_expert} illustrates this gap: a VLM produces generic praise for a Ming-dynasty farewell painting, while an expert critique identifies specific brushstroke techniques, historical conventions, and aesthetic principles.\looseness=-1

% --- Move 2: Establishing the Niche ---

Existing evaluation fails to capture this gap for two reasons. First, benchmarks assess perception without interpretation: POPE~\cite{li2023pope}, VQAv2~\cite{goyal2017vqa}, MME~\cite{fu2025mme}, and SEED-Bench~\cite{li2024seedbench} test whether VLMs see objects (L1) but not whether they understand symbolism (L3) or philosophy (L5). Second, evaluation methods themselves are unreliable: dual-judge averaging suffers from systematic disagreement (we observe cross-judge ICC(2,1)~$= -0.50$), and automated metrics correlate only weakly with human judgement. Recent single-culture studies~\cite{yu-etal-2025-structured,yu-zhao-2025-seeing} find significant VLM-expert divergence on Chinese painting critique, but cannot disentangle culture-specific challenges from systematic Western bias. The central challenge is therefore not only \textit{what} to measure but \textit{how} to measure it reliably.\looseness=-1

% --- Move 3: Occupying the Niche ---

We address this measurement gap with a tri-tier evaluation framework. Tier~I computes automated quality indicators that detect whether relevant cultural concepts are mentioned. Tier~II applies single-judge rubric scoring that additionally assesses whether cultural properties are correctly attributed and interpreted. Tier~III calibrates the aggregate score to human expert ratings via sigmoid calibration. We refer to evaluation stages as ``Tiers'' and to cultural understanding depth as ``Levels'' (L1--L5) throughout.\looseness=-1

This paper is organised around three research questions:

\begin{enumerate}[label=\textbf{RQ\arabic*}., leftmargin=2.5em, itemsep=1pt]
    \item \textbf{Measurement.} How should VLM cultural understanding be measured beyond perception-level benchmarks?
    \item \textbf{Reliability.} How reliable are current evaluation proxies for cultural interpretation?
    \item \textbf{Diagnosis.} Do VLMs exhibit systematic cultural understanding gaps and biases?
\end{enumerate}

Our contributions, each a verifiable answer to the corresponding question:

\begin{enumerate}[label=\textbf{C\arabic*}., leftmargin=2.5em, itemsep=1pt]
    \item \textbf{Construct definition} (RQ1): a tri-tier evaluation framework with 165 culture-specific dimensions across six traditions, anchored to expert judgement.
    \item \textbf{Measurement validation} (RQ2): empirical evidence that dual-judge averaging and automated proxies are unreliable for cultural evaluation, establishing single-judge sigmoid calibration as the reliable alternative.
    \item \textbf{Diagnostic findings} (RQ3): applying the validated instrument to 15 VLMs across six cultural traditions, we find that cultural understanding degrades from L1--L2 to L3--L5 (Depth ($\sigma{=}0.56$) and Alignment ($\sigma{=}0.48$) most discriminative), with 13 of 15 models assigning higher scores to Western art (Cohen's $d = -0.74$, $p < 0.001$).
\end{enumerate}

% =============================================================================
% End of Introduction Section
% =============================================================================

% Section 2: Related Works
% =============================================================================
% Paper 1: Related Works Section (Compressed per D5)
% =============================================================================

\section{Related Work}
\label{sec:related}

\subsection{VLM Evaluation Benchmarks}

VLM benchmarks have progressed from perceptual grounding (VQAv2~\cite{goyal2017vqa}, GQA~\cite{hudson2019gqa}, POPE~\cite{li2023pope}) to multi-task evaluation (MME~\cite{fu2025mme}, SEED-Bench~\cite{li2024seedbench}), but assess L1--L2 perception rather than cultural interpretation. Third-generation cultural probes such as CulturalBench~\cite{chiu2025culturalbench} (text-only QA, 45 regions), CulturalVQA~\cite{nayak2024culturalvqa} (VLMs, 11 countries), and GIMMICK~\cite{schneider2025gimmick} (144 countries) test factual cultural knowledge but not generative critique, and none span the full L1--L5 hierarchy.\looseness=-1

\subsection{Art Interpretation Theory and Computational Approaches}

Our framework draws on two complementary traditions in the philosophy of art. \citeauthor{panofsky1939}'s~(\citeyear{panofsky1939}) iconological method distinguishes pre-iconographic description (recognising pure forms and objects), iconographic analysis (identifying conventional themes and symbols), and iconological interpretation (grasping intrinsic cultural and philosophical meaning). We operationalise this ascending hierarchy---from perception through convention to interpretation---as five levels that isolate empirically separable VLM competencies. L1 (visual perception) captures the sensory recognition at the core of Panofsky's pre-iconographic stratum, while L2 (technical analysis) foregrounds medium and technique identification that practical observation presupposes yet that demands domain training. L3 (cultural symbolism) targets the conventional subject-matter decoding of iconographic analysis, while L4 (historical context) isolates art-historical attribution, which Panofsky distributes across his second and third strata. L5 (philosophical aesthetics) corresponds to iconological interpretation. \citeauthor{goodman1968}'s~(\citeyear{goodman1968}) theory of symbolic reference motivates the tier distinction: Tier~I targets denotational signals and Tier~II additionally assesses exemplification and expression.\looseness=-1

Computationally, style classification (WikiArt~\cite{saleh2016wikiart}, OmniArt~\cite{strezoski2018omniart}) and affective computing (ArtEmis~\cite{achlioptas2021artemis}) address L1--L2; recent VLM approaches target art history comprehension~\cite{arthistory2024} and formal visual analysis~\cite{bin2024gallerygpt} but not cultural interpretation. \citet{zhang2024computational} survey computational approaches to Chinese painting through Xie He's Six Principles (c.~550) but find implementations limited to L1--L2, and \citet{jing2023yijing} theorises \textit{yijing} without VLM evaluation. Our framework extends to L3--L5 through culture-specific dimensions (\S\ref{sec:dataset}).\looseness=-1

\subsection{Cultural AI and Bias Studies}

Structured knowledge resources such as WuMKG~\cite{wan2024wumkg} for Chinese painting and calligraphy and CulTi~\cite{yuan2025culti} for cultural heritage retrieval enable factual lookup but not generative critique. Bias studies find Western cultural bias in LLM outputs~\cite{tao2024cultural} and fairness disparities across demographic attributes in VLMs~\cite{wu2025fairness}, though at survey level rather than depth of understanding. Prior single-culture VLM art critique work includes \citet{yu-etal-2025-structured}, who found VLM-expert divergence on \textit{qiyun shengdong} (spirit resonance), and \citet{yu-zhao-2025-seeing}, who documented symbolic shortcuts on non-Western fire traditions. However, single-culture designs cannot disentangle culture-specific challenges from systematic bias, motivating our cross-cultural approach.\looseness=-1

\paragraph{Research Gap.}
LLM-as-judge approaches (G-Eval~\cite{liu2023geval}, MT-Bench~\cite{zheng2023judging}, Prometheus-Vision~\cite{lee2024prometheus}) provide scalable evaluation but face three challenges in cultural critique: strong cross-judge disagreement (\S\ref{subsec:judge_selection}), absent culture-specific vocabularies, and Western-dominant training distribution. Table~\ref{tab:related_comparison} situates our work: we uniquely combine cross-cultural scope (six traditions, 165 dimensions), full L1--L5 evaluation, and human-calibrated scoring. In sum, prior work probes factual cultural knowledge through closed-form recognition. Our framework instead targets open-ended generative critique and provides measurement validity through human calibration.

\begin{table}[t]
\centering
\footnotesize
\begin{tabular}{lccc}
\toprule
\textbf{Work} & \textbf{Cultures} & \textbf{Levels} & \textbf{Validation} \\
\midrule
MME/SEED & General & L1 & Auto only \\
CulturalVQA & 11 & L3 (VQA) & Auto only \\
GIMMICK & 144 & L3 (recog.) & Auto only \\
\citet{yu-etal-2025-structured} & Chinese & L1--L5 & Expert \\ \midrule
\textbf{Ours} & 6 & L1--L5 & Tri-Tier \\
\bottomrule
\end{tabular}
\caption{Comparison with prior VLM cultural evaluation work.}
\label{tab:related_comparison}
\end{table}

% =============================================================================
% End of Related Works Section
% =============================================================================

% Section 3: Methodology
% =============================================================================
% Paper 1: Methodology Section (Compressed)
% =============================================================================

\section{Methodology}
\label{sec:methodology}

\begin{figure*}[t]
\centering
\resizebox{0.70\textwidth}{!}{% VULCA-Bench Evaluation Framework - TikZ Version v3.2
% Design: L1-L5 center, colored span bars, grouped backgrounds
% v3.2 fixes: bar alignment, vertical compaction, global labels, header spacing

% Define colors
\definecolor{inputcolor}{RGB}{240,240,240}
\definecolor{tier1color}{RGB}{100,150,200}
\definecolor{tier1bg}{RGB}{225,235,248}
\definecolor{tier2color}{RGB}{80,160,80}
\definecolor{tier2bg}{RGB}{225,245,225}
\definecolor{tier3color}{RGB}{200,140,60}
\definecolor{tier3bg}{RGB}{255,240,215}
\definecolor{outputcolor}{RGB}{255,245,220}
\definecolor{l1color}{RGB}{173,216,230}
\definecolor{l2color}{RGB}{152,200,180}
\definecolor{l3color}{RGB}{255,235,150}
\definecolor{l4color}{RGB}{255,200,150}
\definecolor{l5color}{RGB}{220,160,160}
\definecolor{surfline}{RGB}{130,130,130}

\begin{tikzpicture}[
    levelbox/.style={rectangle, rounded corners=2pt, draw=gray!60, minimum height=0.6cm, align=center, font=\small, inner sep=3pt},
    mainbox/.style={rectangle, rounded corners=3pt, draw, minimum height=0.6cm, align=center, font=\small},
    metriclabel/.style={font=\scriptsize\bfseries, align=center},
    arrow/.style={-{Stealth[length=2.5mm]}, thick},
]

% ============================================================
% BACKGROUND REGIONS (drawn first, behind everything)
% ============================================================
\begin{scope}[on background layer]
    % Tier I background region (left)
    \fill[tier1bg, rounded corners=6pt] (-4.85, -0.45) rectangle (-1.85, 3.6);
    % Tier II background region (right)
    \fill[tier2bg, rounded corners=6pt] (1.85, -0.45) rectangle (4.85, 3.6);
\end{scope}

% ============================================================
% CENTER: L1-L5 Cultural Understanding Hierarchy
% ============================================================
\node[levelbox, fill=l1color, minimum width=3.4cm] (l1) at (0, 0)    {\textbf{L1} Visual};
\node[levelbox, fill=l2color, minimum width=3.2cm] (l2) at (0, 0.75) {\textbf{L2} Technique};
\node[levelbox, fill=l3color, minimum width=3.0cm] (l3) at (0, 1.5)  {\textbf{L3} Symbolism};
\node[levelbox, fill=l4color, minimum width=2.8cm] (l4) at (0, 2.25) {\textbf{L4} Context};
\node[levelbox, fill=l5color, minimum width=2.6cm] (l5) at (0, 3.0)  {\textbf{L5} Philosophy};

% Surface/Deep boundary (prominent)
\draw[surfline, thick, densely dashed] (-4.8, 1.125) -- (4.8, 1.125);
\node[font=\tiny\scshape, surfline, fill=white, inner sep=2pt] at (0, 1.125) {surface \textbar{} deep};

% ============================================================
% LEFT: Tier I — Colored Span Bars
% ============================================================

% Tier I header
\node[font=\small\bfseries, tier1color] at (-3.35, 3.4) {Tier~I: Auto Metrics};

% --- Bar 1: DCR, CSA → L1-L5 (full span) ---
\fill[tier1color!25, rounded corners=2pt] (-2.5, -0.3) rectangle (-2.1, 3.3);
\draw[tier1color!60, rounded corners=2pt] (-2.5, -0.3) rectangle (-2.1, 3.3);
\node[metriclabel, tier1color, rotate=90] at (-2.3, 1.5) {DCR, CSA};

% --- Bar 2: CDS → L1-L5 with L3-L5 emphasis ---
% L1-L2 portion (lighter, dashed)
\fill[tier1color!12, rounded corners=2pt] (-3.25, -0.3) rectangle (-2.85, 1.125);
\draw[tier1color!40, rounded corners=2pt, densely dashed] (-3.25, -0.3) rectangle (-2.85, 1.125);
% L3-L5 portion (darker, emphasized)
\fill[tier1color!35, rounded corners=2pt] (-3.25, 1.125) rectangle (-2.85, 3.3);
\draw[tier1color!70, rounded corners=2pt] (-3.25, 1.125) rectangle (-2.85, 3.3);
\node[metriclabel, tier1color, rotate=90] at (-3.05, 2.2) {CDS};
\node[font=\tiny, tier1color!60] at (-3.05, 0.4) {$\downarrow$};

% --- Bar 3: LQS → level-agnostic (dotted) ---
\fill[tier1color!8, rounded corners=2pt] (-4.0, -0.3) rectangle (-3.6, 3.3);
\draw[tier1color!30, rounded corners=2pt, densely dotted] (-4.0, -0.3) rectangle (-3.6, 3.3);
\node[metriclabel, tier1color!60, rotate=90] at (-3.8, 1.5) {LQS};

% ============================================================
% RIGHT: Tier II — Colored Span Bars
% ============================================================

% Tier II header
\node[font=\small\bfseries, tier2color] at (3.35, 3.4) {Tier~II: LLM Judge};

% --- Bar 4: Cov, Acc → L1-L5 (full span) ---
\fill[tier2color!20, rounded corners=2pt] (2.1, -0.3) rectangle (2.5, 3.3);
\draw[tier2color!50, rounded corners=2pt] (2.1, -0.3) rectangle (2.5, 3.3);
\node[metriclabel, tier2color, rotate=90] at (2.3, 1.5) {Cov, Acc};

% --- Bar 5: Depth, Align → L3-L5 (aligned with surface/deep boundary) ---
\fill[tier2color!30, rounded corners=2pt] (2.85, 1.125) rectangle (3.25, 3.3);
\draw[tier2color!60, rounded corners=2pt] (2.85, 1.125) rectangle (3.25, 3.3);
\node[metriclabel, tier2color, rotate=90] at (3.05, 2.2) {Depth, Align};

% --- Bar 6: Qual → level-agnostic (dotted) ---
\fill[tier2color!8, rounded corners=2pt] (3.6, -0.3) rectangle (4.0, 3.3);
\draw[tier2color!30, rounded corners=2pt, densely dotted] (3.6, -0.3) rectangle (4.0, 3.3);
\node[metriclabel, tier2color!60, rotate=90] at (3.8, 1.5) {Qual};

% ============================================================
% TOP: Tier III + Output
% ============================================================
\node[mainbox, fill=tier3bg, draw=tier3color, thick, minimum width=4.0cm, minimum height=0.6cm]
    (tier3) at (0, 4.35) {\textbf{Tier~III:} Sigmoid Calibration $g^*(\cdot)$};

\node[mainbox, fill=outputcolor, draw=tier3color, very thick, minimum width=3.6cm, minimum height=0.6cm]
    (output) at (0, 5.1) {$S_{\text{II}}^* = g^*(S_{\text{II}})$};

% ============================================================
% BOTTOM: Input
% ============================================================
\node[mainbox, fill=inputcolor, draw=gray, minimum width=4.0cm, minimum height=0.6cm]
    (input) at (0, -1.15) {Input: VLM Critique $c$};

% ============================================================
% ARROWS (Flow)
% ============================================================

% Input → L1 (critique enters assessment)
\draw[arrow, gray] (input.north) -- (l1.south);

% Tier II → Tier III (primary path, thick green, from Tier II bg center)
\draw[arrow, tier2color, very thick] (3.35, 3.65) -- (1.2, 4.02);
\node[font=\tiny, tier2color, right, inner sep=1pt] at (2.5, 3.75) {$S_{\text{II}}$};

% Tier I → Tier III (supplementary, dashed blue, from Tier I bg center)
\draw[arrow, tier1color, dashed, thick] (-3.35, 3.65) -- (-1.2, 4.02);
\node[font=\tiny, tier1color, left, inner sep=1pt] at (-2.5, 3.75) {$S_{\text{I}}$};

% Tier III → Output
\draw[arrow, tier3color, very thick] (tier3.north) -- (output.south);

% ============================================================
% LEGEND (minimal: only bar fill patterns, other cues self-labeled)
% ============================================================
\node[font=\scriptsize, align=center] at (0, -1.85) {%
    \colorbox{tier1color!25}{\rule{0pt}{6pt}\rule{0.5cm}{0pt}}\,%
    \colorbox{tier2color!20}{\rule{0pt}{6pt}\rule{0.5cm}{0pt}}~L1--L5 span
    \qquad
    \colorbox{tier1color!35}{\rule{0pt}{6pt}\rule{0.5cm}{0pt}}\,%
    \colorbox{tier2color!30}{\rule{0pt}{6pt}\rule{0.5cm}{0pt}}~L3--L5 span
    \qquad
    \colorbox{tier1color!8}{\rule{0pt}{6pt}\rule{0.5cm}{0pt}}\,%
    \colorbox{tier2color!8}{\rule{0pt}{6pt}\rule{0.5cm}{0pt}}~Level-agnostic
};

\end{tikzpicture}}
\caption{\textsc{Vulca-Bench} Evaluation Framework. The L1--L5 cultural understanding hierarchy (centre) is the construct being measured. Coloured span bars show which levels each metric assesses: DCR, CSA, Coverage, and Accuracy span L1--L5 (full bars), while CDS, Depth, and Alignment target L3--L5 (darker bars). LQS and Quality are level-agnostic (dotted). The dashed line marks the surface/deep boundary. The Tier~II aggregate is calibrated to human ratings via Tier~III, producing $S_{\text{II}}^*$.}
\label{fig:framework}
\end{figure*}
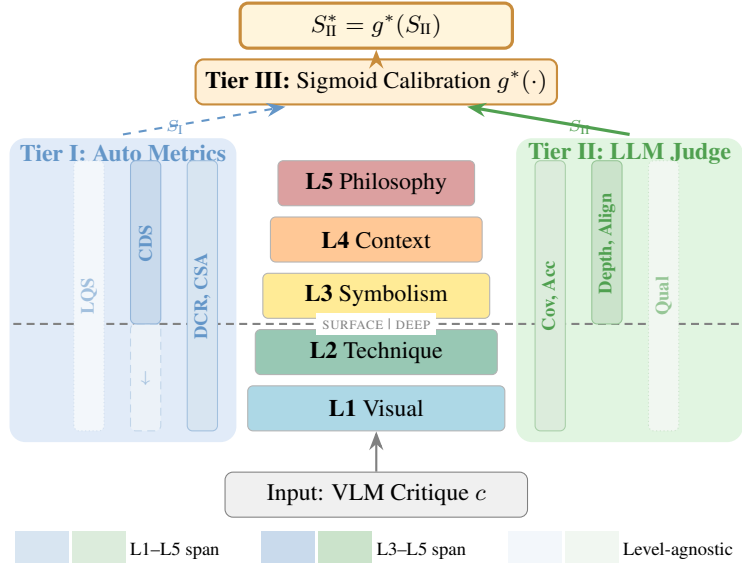

We propose a tri-tier evaluation methodology for cross-cultural art critique assessment (see Figure~\ref{fig:framework}). Following the theoretical grounding in \S\ref{sec:related}, the design reflects the key distinction between surface-level lexical signals and deeper cultural understanding: automated metrics capture keyword presence, while judge-based scoring assesses whether cultural concepts are correctly attributed and interpreted. The three tiers combine automated metrics, single-judge evaluation, and human-focused calibration to address this distinction systematically.

\subsection{Tri-Tier Framework}
\label{subsec:tri_tier}

We distinguish \textit{Levels} (L1--L5) as the construct under measurement from \textit{Tiers} (I--III) as the measurement mechanisms applied to that construct (see Figure~\ref{fig:framework} and Table~\ref{tab:tier_level_map}). The evaluation framework processes a VLM-generated critique $c$ for culture $k$ through three tiers. Tier I and Tier II operate in parallel: Tier~I computes four automated quality indicators, while Tier II produces rubric-based scores on five different cultural dimensions from a single primary judge. Then, Tier III calibrates the Tier II aggregate score to human expert ratings via sigmoid calibration.

\paragraph{Input/Output.}

The framework receives as input (i) a bilingual (Chinese--English) VLM-generated critique asked to cover L1--L5 levels of cultural understanding for a specific artwork image, (ii) a culture tag, and (iii) an expert reference critique for the same artwork. Bilingual output is required because Chinese preserves culture-specific terminology essential for keyword-based Tier~I metrics (e.g., \textit{qiyun} \begin{CJK}{UTF8}{gbsn}气韵\end{CJK}), while English ensures cross-cultural accessibility. Tier~I metrics use only (i) and (ii). Tier~II operates in \textit{reference-guided} mode, where the judge compares the VLM critique against the expert reference to assess cultural alignment and factual accuracy. The framework produces as output (a) the calibrated aggregate score ($S_{\text{II}}^*$), (b) a rubric-based assessment covering five dimensions, and (c) automated risk indicators flagging potential failure types (e.g., low cultural coverage, weak semantic alignment, or elevated template risk).

\paragraph{Tier I: Automated Metrics.}
Tier~I captures surface-level lexical signals: it detects whether a critique \textit{mentions} relevant cultural concepts through keyword and vocabulary matching, without assessing whether those concepts are correctly attributed or interpreted. While these heuristics cannot replace expert perspective, they provide a quantitative signal about potential limitations. We define four automated metrics for a VLM-generated critique $c$ and culture $k$.

%Tier~I functions as a heuristic risk indicator rather than a primary semantic evaluator. While not intended to serve as a standalone substitute for judgement, it provides a quantitative signal used specifically within the optional robustness metric $S_{\text{robust}}$ to penalise fluent-but-shallow templates. Let $c$ be a VLM-generated critique for culture $k$ with dimension set $\mathcal{D}_k$:

The Dimension Coverage Ratio (DCR, Eq.~\ref{eq:dcr}) measures the proportion of culture-specific dimensions $\mathcal{D}_k$ mentioned in $c$:
\begin{equation}
\text{DCR}_\text{auto}(c, k) = \frac{|\mathcal{D}_k^c|}{|\mathcal{D}_k|}
\label{eq:dcr}
\end{equation}
where $\mathcal{D}_k^c \subseteq \mathcal{D}_k$ are dimensions detected via keyword matching (e.g., $|\mathcal{D}_{\text{Chinese}}| = 30$). %Validation on 294 expert anchors yields $r = 0.68$ correlation with human coverage ratings.

The Cultural Semantic Alignment (CSA, Eq.~\ref{eq:csa}) measures TF-IDF cosine similarity with culture-specific vocabulary:

\begin{equation}
\text{CSA}_\text{auto}(c, k) = \cos(\mathbf{v}_c, \mathbf{v}_k) \label{eq:csa}
\end{equation}

where $\mathbf{v}_c$ and $\mathbf{v}_k$ are the TF-IDF vector representations of the critique and the culture-specific vocabulary respectively.

The Critique Depth Score (CDS, Eq.~\ref{eq:cds}) measures whether a critique addresses deeper cultural understanding levels. We assign increasing weights to higher levels, as L3--L5 (symbolism, context, philosophy) require genuine cultural insight rather than surface description:

\begin{equation}
\text{CDS}_\text{auto}(c, k) = \sum_{\ell=1}^{5} w_\ell \cdot \mathbb{1}[\text{L}\ell\text{ covered}] \label{eq:cds}
\end{equation}

where $\mathbb{1}[\cdot]$ is the indicator function (1 if level $\ell$ is detected via keyword matching, 0 otherwise), and $w_\ell = \ell/15$ assigns weights L1:1/15, L2:2/15, \ldots, L5:5/15.

Finally, the Linguistic Quality Score (LQS, Eq.~\ref{eq:lqs}) detects fluent-but-shallow responses by combining length adequacy with sentence complexity:

\begin{equation}
\text{LQS}_\text{auto}(c, k) = \min\left(1, \frac{|c|}{L_{\max}}\right) \cdot \frac{n_{\text{sent}}}{n_{\text{sent}} +
\epsilon} \label{eq:lqs}
\end{equation}

where $|c|$ is the character count, $L_{\max}=2000$ the expected expert-critique length, $n_{\text{sent}}$ the sentence count, and $\epsilon=3$ a smoothing constant.

Each metric is rescaled to $[1,5]$ via $\hat{m}=1+4\cdot\min(1,\max(0,m))$ and the Tier~I aggregate is their mean: $S_{\text{I}}=\frac{1}{4}\sum_{m\in\mathcal{M}_{\text{I}}}\hat{m}$.
Tier~I signals serve as diagnostic risk indicators rather than primary ranking metrics.

\paragraph{Tier II: Judge Scoring.}
Unlike Tier~I, Tier~II assesses deeper cultural understanding: the judge evaluates not only whether concepts are mentioned but whether cultural properties are correctly attributed and whether deeper aesthetic meanings are appropriately interpreted. In \textit{reference-guided} mode (defined above), the judge scores five dimensions: Coverage (L1--L5 breadth), Alignment (culture-specificity), Depth (L3--L5 interpretation quality), Accuracy (factual correctness), and Quality (coherence). Each dimension receives a 1--5 rating, and the aggregate score is their mean:
\begin{equation}
S_{\text{II}} = \frac{1}{5}\sum_{i=1}^{5} d_i
\label{eq:tier2}
\end{equation}
where $d_i$ is the score for dimension $i$.

% \paragraph{L5 Subjectivity.}
%It is important to note that L5 dimensions (philosophical aesthetics) are often interpretive. Concepts such as \textit{qiyun shengdong}, \textit{wabi-sabi}, and \textit{rasa} admit legitimate expert disagreement. To address this variance, we employ isotonic regression~\cite{niculescu2005predicting}. While raw inter-annotator agreement is $\kappa_w = 0.39$ (`fair'), with L3--L5 showing lower agreement ($\kappa = 0.35$) than L1--L2 ($\kappa = 0.52$), the Tier III isotonic calibration partially mitigates these systematic differences.\looseness=-1

%\paragraph{Anchor Handling Across Cultures.}
%In \textit{reference-guided} mode, the judge receives bilingual expert critiques as alignment context. For Chinese art, anchors are in Chinese with English translation; for Western, Japanese, Korean, Islamic, and Indian art, anchors are in English with the original language where available. All anchors are truncated to 1,200 characters per language. The anchor provides semantic alignment context, rather than information leakage: ablations confirm wrong-culture anchors degrade scores (Shuffled~2.14~$\ll$~Matched~3.26, $d{=}{-}1.8$), and removing anchors yields \textit{higher} scores (3.48 vs.~3.26), indicating anchors raise the evaluation bar rather than inducing text-matching.\looseness=-1

%\paragraph{Judge Selection.}
We adopt a single-judge design to ensure same-scale outputs. The judge evaluates each VLM critique against the paired expert reference using the five-dimension rubric, and the aggregate score is then calibrated via Tier~III sigmoid calibration. The framework also supports a \textit{reference-free} mode (rubric-only, no expert anchor) for deployment settings where references are unavailable. All results in this paper use reference-guided evaluation. Crucially, the expert critique serves as an alignment anchor rather than a unique gold answer, since L3--L5 dimensions admit legitimate interpretive plurality.

Table~\ref{tab:tier_level_map} summarises how each evaluation component maps onto the L1--L5 hierarchy, making the construct--mechanism relationship explicit.

\begin{table}[t]
\centering
\footnotesize
\begin{tabular}{lcc}
\toprule
\textbf{Component} & \textbf{L1--L2} & \textbf{L3--L5} \\
\midrule
\multicolumn{3}{l}{\textit{Tier~I (Automated)}} \\
\quad DCR & \checkmark & \checkmark \\
\quad CSA & \checkmark & \checkmark \\
\quad CDS & -- & \checkmark \\
\quad LQS & -- & -- \\
\midrule
\multicolumn{3}{l}{\textit{Tier~II (Judge)}} \\
\quad Coverage & \checkmark & \checkmark \\
\quad Accuracy & \checkmark & \checkmark \\
\quad Depth & -- & \checkmark \\
\quad Alignment & -- & \checkmark \\
\quad Quality & -- & -- \\
\midrule
\multicolumn{3}{l}{\textit{Tier~III}} \\
\quad Calibration & \multicolumn{2}{c}{aggregate only} \\
\bottomrule
\end{tabular}
\caption{Tier--Level mapping. Checkmarks indicate which cultural understanding levels each component explicitly assesses. L1--L2 covers visual perception and technical analysis; L3--L5 covers cultural symbolism, historical context, and philosophical aesthetics.}
\label{tab:tier_level_map}
\end{table}

\paragraph{Agreement Reporting (ICC).}
For agreement analyses, we use \textbf{ICC(2,1)} (two-way random effects, absolute agreement, single measurement) following reporting guidance in Koo \& Li (2016) and McGraw \& Wong (1996)~\cite{koo2016guideline,mcgraw1996forming}. We report 95\% confidence intervals for Tier~I--Tier~II ICC estimates (Appendix Table~\ref{tab:icc_full}). We use ICC rather than Pearson/Spearman correlation because our target is \emph{interchangeability under a shared scale} (absolute agreement), while correlation measures association and can remain high under additive or scaling mismatches.

\paragraph{Tier III: Human Calibration.}
Tier~III grounds the evaluation in human expert judgement, addressing the concern that even judge-based scoring may exhibit systematic biases. Calibration aligns $S_{\text{II}}$ (Eq.~\ref{eq:tier2}) with human ratings via sigmoid calibration~\cite{niculescu2005predicting}:\looseness=-1

\begin{equation}
S_{\text{II}}^* = 1 + 4\,\sigma(a \cdot S_{\text{II}} + b)
\label{eq:calibration}
\end{equation}

where $\sigma(\cdot)$ is the logistic function and $a, b$ are fitted by minimising MSE on the training split of human-scored pairs ($n=295$), with $S_{\text{h}}$ denoting the human score. The sigmoid maps the unbounded judge aggregate onto $[1,5]$ while preserving monotonicity by construction. Dimension-specific scores remain uncalibrated (Tier~II) but are reported for analytical purposes. Beyond reducing absolute error, calibration enables comparability across models and cultures by anchoring rankings to an explicit measurement scale grounded in human expertise, a requirement for cross-cultural claims where systematic bias is itself under investigation.

%\paragraph{Primary vs Robust Scoring.}
%The primary score $S_{\text{primary}} = S_{\text{II}}^*$\label{eq:primary} maximises agreement with human ratings. For adversarial settings (e.g., fluent-but-shallow templates), we report $S_{\text{robust}} = 0.4 \cdot S_{\text{I}} + 0.6 \cdot S_{\text{II}}^*$\label{eq:robust}; weights in $[0.3, 0.5]$ achieve $\rho \geq 0.97$ with human judgements.

\section{Dataset}
\label{sec:dataset}

To validate our framework, we use the \textsc{Vulca-Bench} corpus \cite{yu2026vulca},\footnote{Dataset: \url{\dataseturl}} our companion dataset paper that details the full annotation protocol and quality assurance process. The complete corpus contains 6,804 matched pairs across 7 cultural traditions. However, we exclude the Mural tradition (109 pairs) from evaluation due to an insufficient number of expert annotations for reliable Tier III calibration, resulting in a 6-culture evaluation subset of 6,695 image-critique pairs covering a total of 165 culture-specific dimensions (see Appendix~\ref{app:dimensions} for a detailed list of the dimensions). From this subset, we extract 294 stratified samples with bilingual expert-written critiques which we use as evaluation samples (Eval.) for our experiments. On top of that, a disjoint 450-sample human-scored (HScored) subset of art critiques (295 training / 155 held-out test) serves as the gold-standard $S_{\text{h}}$ for fitting and validating the sigmoid calibration function $g^*$ in Tier~III. The pool size was determined by available annotator capacity; the 65/35 train-test split balances calibration stability against held-out evaluation power (see Appendix~\ref{app:calibration_baseline} for a sample-size sensitivity analysis). See Table \ref{tab:dataset_stats} for a complete overview of our data distribution.

Each image-critique pair consists of (i) an artwork image, (ii) artist and title metadata, and (iii) an expert bilingual critique (Chinese: $\sim$500 characters; English: $\sim$300 words) annotated with culture-specific dimensions. Critiques explicitly identify L1--L5 layer coverage through dimension-specific tags (e.g., CN\_L1\_D1 for Chinese Visual Colour, see Appendix~\ref{app:dimensions}), enabling the automatic computation of cultural coverage metrics.

\begin{table}[t]
\centering
\footnotesize
\begin{tabular}{lrrrrr}
\toprule
\textbf{Culture} & \textbf{Dims} & \textbf{Pairs} & \textbf{Eval.} & \textbf{HScored} \\
\midrule
Chinese & 30 & 1,854 & 50 & 99 \\
Western & 25 & 4,270 & 50 & 97 \\
Japanese & 27 & 201 & 46 & 84 \\
Korean & 25 & 107 & 48 & 48 \\
Islamic & 28 & 165 & 50 & 52 \\
Indian & 30 & 98 & 50 & 70 \\
\midrule
\textbf{Total} & \textbf{165} & \textbf{6,695} & \textbf{294} & \textbf{450} \\
\bottomrule
\end{tabular}
\caption{Dataset distribution. Eval. column contains the sample distribution used for evaluating VLMs. HScored describes the distribution of human-scored art critiques.} 
\label{tab:dataset_stats}
\end{table}

\paragraph{Human-Scored Art Critiques.}
The 450-sample human-scored subset was annotated by domain experts via a balanced incomplete block design: three annotator pairs each scored ${\sim}100$ art critiques independently on the five Tier~II dimensions (1--5 scale, 0.5-step increments), yielding 299 dual-rated items for agreement analysis. Overall inter-annotator agreement was moderate ($\kappa_w = 0.43$, 95\% CI $[0.39, 0.45]$; \citealp{landis1977measurement}). Surface-oriented dimensions (Coverage, Quality) showed higher agreement ($\kappa_w = 0.54$ $[0.48, 0.58]$) than L3--L5 dimensions requiring cultural interpretation (Alignment, Depth, Accuracy: $\kappa_w = 0.35$ $[0.31, 0.39]$), evidencing the inherent subjectivity of deeper cultural understanding assessment.\footnote{These $\kappa_w$ values measure agreement on \textit{ordinal rubric ratings} (1--5 scale on Tier~II dimensions), a harder task than the binary dimension-presence agreement ($\kappa = 0.77$) reported in \citet{yu2026vulca}. The two figures are not directly comparable.}

% =============================================================================
% End of Methodology Section
% =============================================================================

% Section 4: Experimental Setup
% =============================================================================
% Paper 1: Experiments Section (Compressed)
% =============================================================================

\section{Experiments}
\label{sec:experiments}

%We evaluate 15 VLMs on 294 expert anchors, conducting 4,406 evaluations after filtering 4 instances\footnote{Filtered instances: 2 safety refusals (DeepSeek-VL2 on religious iconography), 2 malformed JSON (GLM-4V-Flash timeout errors). All affected models had $\geq$290 successful evaluations; no systematic bias detected.}.

\subsection{Experimental Setup}
\label{subsec:setup}

We evaluate 15 VLMs on art critique generation using 294 evaluation samples spanning six cultural traditions, yielding 4,405 model--sample evaluations.\footnote{15 models $\times$ 294 samples = 4,410. Five instances (0.11\%) were excluded: 4 malformed JSON responses (GLM-4V-Flash, 1.4\% of its evaluations) and 1 empty critique (GPT-5, 0.3\%). All remaining 13 models had zero failures. Imputing the excluded instances with a default score of 3.0 instead of excluding them does not change any model ranking.} The models cover five providers (both closed and open weight): GPT-5, GPT-5.2, GPT-5-mini, GPT-4o, GPT-4o-mini, Claude-Sonnet-4.5, Claude-Opus-4.5, Claude-Haiku-3.5, Gemini-2.5-Pro, Gemini-3-Pro, Qwen3-VL-235B, Qwen3-VL-Flash, Llama4-Scout, GLM-4V-Flash, and DeepSeek-VL2.

Each model receives a compressed artwork image ($\leq$3.75MB) and generates a bilingual L1--L5 critique using a unified prompt. Tier~I automated metrics are computed on the generated critique. Claude Opus 4.5 (Tier~II judge) then scores the five rubric dimensions in \textit{reference-guided} mode (\S\ref{subsec:tri_tier}). The aggregate Tier~II score is calibrated via sigmoid calibration on the human-scored subset (\S\ref{sec:dataset}). Our default evaluation metric is the calibrated score $S_{\text{II}}^*$; dimension-specific Tier~II scores are reported for diagnostic purposes.

%; $S_{\text{robust}}$ is reported as a supplementary robustness signal and is not used for the main evaluation ordering. We additionally report $S_{\text{robust}} = 0.4\cdot S_{\text{I}} + 0.6\cdot S_{\text{II}}^*$ for settings with formulaic responses, where $S_{\text{I}}$ is the rescaled Tier~I score. All main results use \textit{reference-guided} mode (with expert anchor).

\subsection{Judge Selection}
\label{subsec:judge_selection}

\begin{table}[t]
\centering
\footnotesize
\resizebox{\columnwidth}{!}{%
\begin{tabular}{llcccc}
\toprule
\textbf{Model} & \textbf{Provider} & \textbf{Mean} & \textbf{Std} & \textbf{Lat.} & \textbf{Tendency} \\
\midrule
GPT-4o & OpenAI & 4.52 & 0.17 & 2.9s & Lenient \\
GPT-4o-mini & OpenAI & 4.31 & 0.16 & 1.6s & Lenient \\
GPT-5 & OpenAI & 4.09 & 0.23 & 18.8s & Moderate \\
Claude-Sonnet-4.5 & Anthropic & 3.99 & 0.16 & 14.5s & Moderate \\
GPT-5.2 & OpenAI & 3.76 & 0.16 & 1.7s & Strict \\
Claude-Haiku-3.5 & Anthropic & 3.76 & 0.37 & 7.6s & High Var. \\
GPT-5-mini & OpenAI & 3.65 & 0.31 & 9.0s & Strict \\
\textbf{Claude-Opus-4.5} & Anthropic & \textbf{3.42} & 0.18 & 14.9s & \textbf{Strictest} \\
\bottomrule
\end{tabular}%
}
\caption{Single-judge comparison.}
\label{tab:single_judge_full}
\end{table}

% \paragraph{Why single-judge?}
To select our Tier II judge, we compared 8 different models. As shown in Table \ref{tab:single_judge_full}, OpenAI models provided higher scores (mean 4.1-4.5) while Claude models were stricter (3.4-4.0). Furthermore, we also explored a dual-judge configuration, observing unreliable results where judges exhibited systematic scale mismatches. More specifically, we observed strong ICC(2,1) disagreement in cross-judge scenarios ranging from $-0.50$ (Claude-Opus-4.5 + GPT-5, $n=150$) to $0.12$ (Claude-Sonnet + GPT-5, $n=150$), all below our 0.6 threshold. In the end, single-judge design with sigmoid calibration yielded stable scores aligned with human judgements. Therefore, we selected Claude Opus 4.5 for three main reasons: (1) stable rank discrimination, (2) consistent cultural gap direction, and (3) absence of self-favouritism (i.e., it does not favour its own outputs).

%Cross-judge ICC $= -0.50$ (Claude-Opus-4.5 vs GPT-5, $n=150$) confirms incompatible scales; averaging such scores produces composites that lack meaningful interpretation without explicit scale alignment. 

%\paragraph{Robustness validation.}
%Furthermore, to verify that our findings are not judge-specific, we replicate model-level ranking directionality with an open-source alternative (GLM-4V-Flash, $n=290$ shared evaluations). We observe moderate agreement with the primary judge (Kendall's $\tau = 0.401$, $p<0.001$); see Table~\ref{tab:opensrc_judge}. We treat this as a directional robustness check rather than justification for multi-judge averaging. Claude Opus 4.5 remains our reference implementation; the framework is judge-agnostic via the Tier~III calibration protocol.

\subsection{Calibration Validation}
\label{subsec:calibration_validation}

\begin{table}[t]
\centering
\begin{tabular}{lcc}
\toprule
\textbf{Score} & \textbf{MAE} & \textbf{MAE (cal)} \\
\midrule
Aggregate $S_{\text{II}}$ & 0.454 & 0.446 \\
\midrule
\textbf{$\Delta$} & -- & \textbf{$-$1.7\%} \\
\bottomrule
\end{tabular}
\caption{Human calibration results (n=155, held-out). Tier~III calibrates only the aggregate Tier~II score.}
\label{tab:kappa_full}
\end{table}

Table \ref{tab:kappa_full} reports held-out human calibration (Tier~III). The sigmoid function (Eq.~\ref{eq:calibration}) is fitted on the training split and applied only to the aggregate Tier~II score on held-out data, yielding a 1.7\% MAE reduction. Dimension-specific scores remain uncalibrated (\S\ref{sec:methodology}). A comparison with isotonic regression and a sample-size sensitivity analysis are reported in Appendix~\ref{app:calibration_baseline}.

%To address residual concerns about item-level contamination, we report cross-fitted calibration using 5-fold isotonic regression where each fold's items are calibrated with a function fit on the remaining 4 folds. Cross-fitted MAE ($0.441$) versus global-fit MAE ($0.423$, 298-split) shows only $+4.3\%$ difference, confirming that item-level leakage contributes minimal calibration gain.

% =============================================================================
% End of Experiments Section
% =============================================================================

% Section 5: Results (Updated 2025-12-15 with Phase 2c real data)
% =============================================================================
% Paper 1: Results & Analysis Section
% =============================================================================

\section{Results and Analysis}
\label{sec:results}

%We evaluate 15 VLMs on 294 gold samples (4,406 evaluations; see \S\ref{sec:experiments} for filtering details) to address three research questions.

% =============================================================================
% RQ3: Cross-Cultural VLM Diagnosis
% =============================================================================

\subsection{Cross-Cultural VLM Diagnosis (RQ3)}
\label{subsec:overall}

% @SOURCE: unified_results.db layer2_scores (Claude Opus 4.5 judge, 294 GS samples)
\begin{table}[t]
\centering
\footnotesize
\resizebox{\columnwidth}{!}{%
\begin{tabular}{@{}lc|ccccc@{}}
\toprule
\textbf{Model} & \textbf{$S_{\text{II}}^*$} & \textbf{Cov} & \textbf{Align} & \textbf{Depth} & \textbf{Acc} & \textbf{Qual} \\
\midrule
\rowcolor{green!15} Gemini-2.5-Pro & \textbf{4.27} & \textbf{4.49} & \textbf{4.26} & 4.38 & 3.56 & \textbf{4.55} \\
\rowcolor{green!15} Qwen3-VL-235B & 4.21 & \textbf{4.49} & 4.10 & \textbf{4.41} & 3.33 & 4.51 \\
\rowcolor{green!15} Claude-Sonnet-4.5 & 4.11 & 4.29 & 4.05 & 4.00 & 3.44 & 4.48 \\
\midrule
Claude-Opus-4.5 & 4.09 & 4.37 & 3.90 & 3.99 & 3.65 & 4.27 \\
GPT-5 & 4.00 & 4.23 & 3.48 & 4.04 & \textbf{3.85} & 4.08 \\
GPT-5.2 & 3.97 & 4.17 & 3.53 & 3.97 & 3.66 & 4.19 \\
Gemini-3-Pro & 3.95 & 4.12 & 3.60 & 3.83 & 3.74 & 4.11 \\
Llama4-Scout & 3.67 & 4.21 & 3.48 & 3.36 & 2.96 & 4.10 \\
GPT-5-mini & 3.64 & 3.87 & 2.98 & 3.48 & 3.67 & 3.97 \\
Claude-Haiku-3.5 & 3.61 & 3.80 & 3.18 & 3.46 & 3.35 & 4.05 \\
\midrule
\rowcolor{red!10} Qwen3-VL-Flash & 3.58 & 3.62 & 3.20 & 3.38 & 3.25 & 4.24 \\
\rowcolor{red!10} GPT-4o & 3.57 & 3.88 & 3.38 & 3.21 & 3.09 & 4.10 \\
\rowcolor{red!10} GPT-4o-mini & 3.24 & 3.76 & 2.94 & 2.93 & 2.90 & \underline{3.76} \\
\rowcolor{red!10} GLM-4V-Flash & 3.15 & 3.57 & 2.75 & 2.71 & 2.99 & 3.94 \\
\rowcolor{red!10} DeepSeek-VL2 & \underline{3.01} & \underline{3.50} & \underline{2.74} & \underline{2.64} & \underline{2.72} & 3.78 \\
\midrule
\textit{Std ($\sigma$)} & --- & 0.33 & 0.48 & 0.56 & 0.35 & 0.24 \\
\bottomrule
\end{tabular}%
}
\caption{VLM $S_{\text{II}}^*$ performance with five Tier~II dimensions (Cov=Coverage, Align=Alignment, Depth=Critique Depth, Acc=Accuracy, Qual=Quality). The best results are in bold, the worst are underlined. $S_{\text{II}}^*$ is computed as $\sigma(\bar{S}_{\text{II}})$, i.e., sigmoid calibration applied to the per-model mean of raw dimension averages. Culture-level analyses (Figure~\ref{fig:tier_gap}b) instead average per-evaluation calibrated scores (difference $< 0.05$).}
\label{tab:vlm_performance}
\end{table}

Having established the instrument's measurement properties (\S\ref{sec:experiments}), we now apply it to diagnose VLM cultural understanding. Table~\ref{tab:vlm_performance} reveals clear performance stratification across cultural depth. Gemini-2.5-Pro (4.27) and Qwen3-VL-235B (4.21) produce the most complete critiques, while Claude models remain competitive (Sonnet 4.11, Opus 4.09). The overall spread of 1.26 points reveals meaningful differences across models. The most discriminative dimension is Critique Depth ($\sigma=0.56$), followed by Cultural Alignment ($\sigma=0.48$) and Factual Accuracy ($\sigma=0.35$), precisely the dimensions that correspond to iconographic and iconological interpretation (L3--L5). DeepSeek-VL2, for instance, scores 3.50 on Coverage but only 2.74 on Alignment and 2.64 on Depth, indicating weaker cultural grounding rather than surface-level fluency deficits. Bootstrap 95\% confidence intervals (10K resamples) show that 6 of 14 adjacent-rank pairs are non-overlapping, while the top tier (\#1--3: $S_{\text{II}}^*$ 4.11--4.27) and bottom tier (\#13--15: 3.01--3.24) are fully separated ($p < 0.001$, permutation test). Mid-table ranks with overlapping CIs should be interpreted as performance bands rather than strict orderings.\looseness=-1

Linguistic Quality shows the least variance ($\sigma=0.24$), confirming that all 15 VLMs produce fluent text and that surface polish is not the bottleneck. The discriminative power lies in Depth and Alignment, which capture L3--L5 cultural understanding where pre-iconographic competence no longer suffices. This is consistent with the theoretical distinction in \S\ref{sec:related}: mentioning relevant concepts is relatively easy, while correctly attributing and interpreting cultural properties is where models diverge.\looseness=-1

% =============================================================================
% RQ2: Measurement Reliability
% =============================================================================

\subsection{Automated vs.\ Judge-Based Measurement (RQ2)}
\label{subsec:tier_gap}

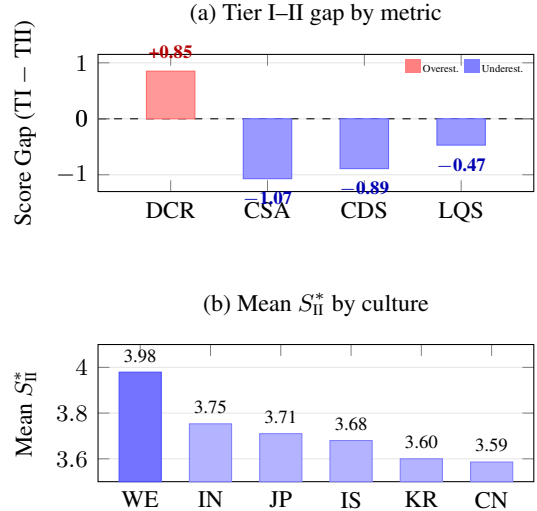
\begin{figure}[t]
\centering
\begin{tikzpicture}
% ===== (a) Tier I--II Gap =====
\begin{axis}[
    name=gapplot,
    width=0.95\columnwidth,
    height=3.4cm,
    ylabel={\footnotesize Score Gap (TI $-$ TII)},
    xtick={0,1,2,3},
    xticklabels={DCR,CSA,CDS,LQS},
    ymin=-1.3, ymax=1.15,
    enlarge x limits=0.25,
    extra y ticks={0},
    extra y tick style={grid style={dashed,black,line width=0.4pt}},
    tick label style={font=\footnotesize},
    label style={font=\footnotesize},
    ymajorgrids=true,
    grid style={gray!20},
    clip=false,
    title style={at={(0.5,1.02)}, anchor=south, font=\small},
    title={(a) Tier I--II gap by metric},
]
% Dummy plot to force axis coordinate initialization
\addplot[draw=none, forget plot] coordinates {(0,0) (3,0)};
% Positive bar (overestimate) - red
\fill[red!40, draw=red!60, line width=0.4pt]
    (axis cs:-0.25,0) rectangle (axis cs:0.25,0.85);
\node[above, font=\scriptsize, text=red!70!black, yshift=1pt] at (axis cs:0,0.85) {\textbf{+0.85}};
% Negative bars (underestimate) - blue
\fill[blue!40, draw=blue!60, line width=0.4pt]
    (axis cs:0.75,0) rectangle (axis cs:1.25,-1.07);
\node[below, font=\scriptsize, text=blue!70!black, yshift=-1pt] at (axis cs:1,-1.07) {\textbf{$-$1.07}};
\fill[blue!40, draw=blue!60, line width=0.4pt]
    (axis cs:1.75,0) rectangle (axis cs:2.25,-0.89);
\node[below, font=\scriptsize, text=blue!70!black, yshift=-1pt] at (axis cs:2,-0.89) {\textbf{$-$0.89}};
\fill[blue!40, draw=blue!60, line width=0.4pt]
    (axis cs:2.75,0) rectangle (axis cs:3.25,-0.47);
\node[below, font=\scriptsize, text=blue!70!black, yshift=-1pt] at (axis cs:3,-0.47) {\textbf{$-$0.47}};
% Mini legend (compact, transparent, \rule for precise height)
\node[anchor=north east, font=\fontsize{4}{4.2}\selectfont, fill=none,
    inner ysep=0pt, inner xsep=0.5pt] at (rel axis cs:0.98,0.97) {%
    {\color{red!50}\rule{2mm}{5pt}}~Overest.\enspace
    {\color{blue!50}\rule{2mm}{5pt}}~Underest.};
\end{axis}
% ===== (b) Cultural Bias =====
\begin{axis}[
    at=(gapplot.below south west),
    anchor=above north west,
    yshift=-18pt,
    width=0.95\columnwidth,
    height=3.4cm,
    ylabel={\footnotesize Mean $S_{\text{II}}^*$},
    xtick={0,1,2,3,4,5},
    xticklabels={WE,IN,JP,IS,KR,CN},
    ymin=3.50, ymax=4.10,
    enlarge x limits=0.12,
    tick label style={font=\footnotesize},
    label style={font=\footnotesize},
    ymajorgrids=true,
    grid style={gray!20},
    clip=false,
    title style={at={(0.5,1.02)}, anchor=south, font=\small},
    title={(b) Mean $S_{\text{II}}^*$ by culture},
]
% Dummy plot to force axis coordinate initialization
\addplot[draw=none, forget plot] coordinates {(0,3.50) (5,3.50)};
% WE bar (highlighted - darker)
\fill[blue!55, draw=blue!70, line width=0.4pt]
    (axis cs:-0.3,3.50) rectangle (axis cs:0.3,3.979);
\node[above, font=\scriptsize] at (axis cs:0,3.979) {3.98};
% Other cultures (lighter)
\fill[blue!30, draw=blue!50, line width=0.4pt]
    (axis cs:0.7,3.50) rectangle (axis cs:1.3,3.753);
\node[above, font=\scriptsize] at (axis cs:1,3.753) {3.75};
\fill[blue!30, draw=blue!50, line width=0.4pt]
    (axis cs:1.7,3.50) rectangle (axis cs:2.3,3.710);
\node[above, font=\scriptsize] at (axis cs:2,3.710) {3.71};
\fill[blue!30, draw=blue!50, line width=0.4pt]
    (axis cs:2.7,3.50) rectangle (axis cs:3.3,3.680);
\node[above, font=\scriptsize] at (axis cs:3,3.680) {3.68};
\fill[blue!30, draw=blue!50, line width=0.4pt]
    (axis cs:3.7,3.50) rectangle (axis cs:4.3,3.600);
\node[above, font=\scriptsize] at (axis cs:4,3.600) {3.60};
\fill[blue!30, draw=blue!50, line width=0.4pt]
    (axis cs:4.7,3.50) rectangle (axis cs:5.3,3.586);
\node[above, font=\scriptsize] at (axis cs:5,3.586) {3.59};
\end{axis}
\end{tikzpicture}
\caption{(a)~Tier~I--II gap: \textcolor{red!60!black}{positive} = Tier~I overestimates, \textcolor{blue!60!black}{negative} = Tier~I underestimates relative to Tier~II. (b)~Average calibrated scores by culture; Western art consistently receives highest scores.}
\label{fig:tier_gap}
\end{figure}

A core validity question is whether automated proxies and judge-based scoring measure the same construct. All four Tier~I automated metrics exhibit poor ICC values with Tier~II judge scores ($<$0.5): DCR$_\text{auto}$~(0.02), CSA$_\text{auto}$~(0.17), CDS$_\text{auto}$~(0.18), LQS$_\text{auto}$~($-$0.17). As shown in Figure~\ref{fig:tier_gap}~(a), DCR$_\text{auto}$ overestimates dimension coverage ($+$0.85) by matching surface-level keywords, while CSA$_\text{auto}$/CDS$_\text{auto}$ underestimate ($-$1.07/$-$0.89) cultural alignment, precisely the dimensions most relevant for cross-cultural depth evaluation.

Table~\ref{tab:tier1_human} compares Tier~I metrics with human-scored art critiques ($n=450$). All metrics show weak-to-moderate correlation ($r \in [0.27, 0.53]$), confirming their role as complementary risk indicators rather than standalone evaluation metrics.

The near-zero ICC for DCR (0.02) confirms that lexical coverage and semantic understanding operate as largely independent dimensions rather than coarse and fine-grained measures of the same construct.

%These results do not invalidate Tier I; rather, they clarify its role. We use Tier I signals as \textbf{risk indicators} (e.g., template-risk / superficial keyword coverage) that complement, but cannot replace, calibrated judged quality.

\begin{table}[t]
\centering
\small
\begin{tabular}{lcc}
\toprule
Metric & Pearson $r$ & Interpretation \\
\midrule
DCR$_\text{auto}$ & 0.53 & Moderate \\
CSA$_\text{auto}$ & 0.44 & Weak-Moderate \\
CDS$_\text{auto}$ & 0.51 & Moderate \\
LQS$_\text{auto}$ & 0.27 & Weak \\
\bottomrule
\end{tabular}
\caption{Tier I against human gold standard (n=450). All with $p < 0.001$.}
\label{tab:tier1_human}
\end{table}

% =============================================================================
% RQ3 (cont.): Cultural Disparities
% =============================================================================

\subsection{Cultural Disparities (RQ3)}
\label{subsec:cultural_bias}

% Figure merged into fig:tier_gap above
The instrument also reveals culture-stratified scoring patterns. Figure~\ref{fig:tier_gap}~(b) shows that Western art critiques consistently receive higher scores in our Tier~II evaluation. The gap between Chinese and Western art scores is $-0.39$ (Cohen's $d=-0.74$, computed over $n_{\text{CN}}{=}750$, $n_{\text{WE}}{=}747$ instance-level evaluations; $p < 0.001$, bootstrap 95\% CI $[-0.44, -0.34]$, 1,000 resamples).

%A mixed-effects model with random intercepts for model confirms significance ($\beta = -0.17$, 95\% CI $[-0.27, -0.08]$); see Figure~\ref{fig:mixed_effects} for the full forest plot with all culture pairs. The effect persists after controlling for critique length ($\beta = -0.15$, $p < 0.001$). Cross-judge validation shows the direction is consistent across judges (Table~\ref{tab:crossjudge_full}).

At the model level, thirteen of fifteen VLMs favour Western art critiques, with GPT-4o-mini showing strongest bias ($\Delta=-1.08$) and GPT-5.2 most neutral ($\Delta=+0.07$). Appendix Figure~\ref{fig:model_gap_stratified} visualizes the model-stratified gap distribution.

\paragraph{Confound controls.} Two analyses rule out major confounds. A \textbf{genre-controlled landscape subset} ($n_{\text{CN}}{=}300$, $n_{\text{WE}}{=}405$) yields a larger gap ($d{=}{-}0.93$, 95\% CI $[-1.09, -0.78]$, $p < 0.001$), exceeding the overall effect. A \textbf{blind-culture pilot} ($n{=}50$, GPT-4o) shows the gap \emph{widens} when the culture tag is removed ($\Delta_{\text{blind}}{=}{-}0.61$ vs.\ $\Delta_{\text{std}}{=}{-}0.54$), ruling out judge bias (Appendix~\ref{app:confound_control}).

%At the model level, 11/15 VLMs favor Western art, with GPT-4o showing strongest bias ($\Delta=-0.52$) and Claude-Opus-4.5 most neutral ($\Delta=0.00$). We interpret this as an observed trend under our sampling and rubric, rather than a definitive claim about intrinsic model bias. Differences in genre distribution and metadata richness across cultures may partially contribute; we therefore recommend matched-subset analyses as a follow-up.

Appendix~\ref{app:robustness} provides additional robustness validation, including a Tier~I--Tier~II ICC analysis confirming that automated metrics and judge scores measure distinct constructs (all ICC $< 0.2$).

% Case Study moved to Discussion section per advisor recommendation
% =============================================================================
% End of Results Section
% =============================================================================

% Section 6: Discussion and Conclusion
% =============================================================================
% Paper 1: Discussion & Conclusion Sections (Enhanced)
% =============================================================================

\section{Discussion}
\label{sec:discussion}

As illustrated by representative case studies (Table~\ref{tab:case_study}, Appendix~\ref{app:case_study}), high linguistic quality can mask limited cultural understanding. The tri-tier framework addresses this by combining automated risk indicators (Tier~I) with rubric-based judge evaluation (Tier~II) and human calibration (Tier~III), providing discriminant features missing in previous dual-judge setups.

\paragraph{Failure pattern analysis.} Case studies (Table~\ref{tab:case_study}, Appendix~\ref{app:case_study}) reveal that high keyword coverage can mask shallow understanding: models cite cultural terms generically (e.g., ``ink technique,'' ``artistic conception'') without specifying how they manifest in the target painting. Successful critiques instead identify specific techniques and contextualise them within art-historical traditions. This pattern confirms that the Tier~I--II gap reflects a structural difference in cultural understanding depth.\looseness=-1

\paragraph{Implications for cultural AI.} The L1--L2 to L3--L5 degradation pattern (\S\ref{subsec:overall}) suggests that VLMs trained on image-caption pairs acquire descriptive competence but lack cultural grounding for deeper interpretation. Furthermore, the low Tier~I--II agreement (ICC~$< 0.5$) confirms that automated metrics cannot substitute for semantic evaluation. These findings carry practical consequences for museums, art education platforms, and cultural heritage projects, all of which risk systematic misinterpretation of non-Western traditions without validated evaluation instruments.\looseness=-1

\section{Conclusion}
\label{sec:conclusion}

We return to the three research questions posed in \S1.

\paragraph{RQ1 (Measurement).} Cultural understanding beyond perception can be measured through a tri-tier protocol combining automated indicators, single-judge rubric scoring across 165 culture-specific dimensions, and human calibration, grounded in the art-theoretical hierarchy defined in \S\ref{sec:related}.

\paragraph{RQ2 (Reliability).} Automated proxies and LLM-judge scoring measure different constructs: Tier~I--II ICC remains below 0.2 for all four metrics, and cross-judge ICC reaches $-0.50$. Sigmoid calibration to human ratings via a single judge provides a more reliable alternative.

\paragraph{RQ3 (Diagnosis).} Under the validated instrument, cultural understanding degrades from L1--L2 to L3--L5, with Depth ($\sigma{=}0.56$) and Alignment ($\sigma{=}0.48$) as the most discriminative dimensions. Thirteen of fifteen models assign higher scores to Western art (Cohen's $d = -0.74$, $p < 0.001$), though cross-judge scale incompatibility precludes definitive multi-judge confirmation.

\paragraph{Broader impact.} Single-judge calibration should be preferred over dual-judge averaging in culturally sensitive domains. Future work will scaffold L1$\to$L5 reasoning and extend coverage to additional traditions.

\section*{Acknowledgements}
We thank Ramon Ruiz-Dolz for initial discussions on project framing, early-stage conceptual feedback, and manuscript polishing during the formative phase of this work.

% =============================================================================
% Limitations Section (ACL 2026 REQUIRED - after Conclusion, before References)
% =============================================================================
% =============================================================================
% Limitations Section (ACL 2026 REQUIRED)
% This section MUST appear after Conclusion and before References
% Does NOT count toward page limit
% =============================================================================

\section*{Limitations}
\label{sec:limitations}

\paragraph{Cultural Coverage.}
Our evaluation covers six cultural traditions (6,695 matched pairs: Western 4,270; Chinese 1,854; Japanese 201; Korean 107; Islamic 165; and Indian 98). Chinese and Western samples dominate the distribution, and this imbalance is reflected in our calibration experiments: the held-out test set contains fewer Korean ($n=16$) and Islamic ($n=18$) samples than Chinese or Western ($n=33$ each), and our analysis shows calibration degrades for Islamic ($-6.5\%$) and Indian ($-6.4\%$). 

\paragraph{Language Coverage.}
Our bilingual (Chinese--English) design (\S\ref{sec:methodology}) reflects practical constraints: expert annotators were only available for this language pair. However, this choice may introduce translation loss for non-Chinese traditions: Japanese (\textit{wabi-sabi}, \textit{mono no aware}), Korean (\textit{jeong}, \textit{heung}), Islamic (Arabic calligraphic concepts), and Indian (\textit{rasa}, \textit{bhava}) art possess native terminologies that English romanization can only capture partially. We mitigated this by including romanized terms with diacritics where possible and training annotators on culture-specific vocabulary. However, extending the framework to native-language critiques (Japanese, Korean, Arabic, Hindi/Sanskrit) represents an important future direction. 

\paragraph{Task Subjectivity.}
The Tri-Tier framework depends on a single judge (Claude Opus 4.5) for Tier~II scoring and on the human-scored subset (\S\ref{sec:dataset}) for sigmoid calibration. Cross-cultural art critique evaluation is inherently challenging because L3--L5 dimensions require interpreting culture-specific concepts where legitimate expert disagreement is expected ($\kappa_w = 0.35$ for L3--L5 vs.\ $0.54$ for surface-oriented dimensions; \S\ref{sec:dataset}). Additionally, judge calls use the provider default temperature ($= 1.0$), so scores are non-deterministic. The JSON-constrained integer rubric limits variability, but uncached reruns may produce minor fluctuations.

\paragraph{Rating Scale Choice.}
Tier~II currently uses a fixed 1--5 rubric to preserve comparability with prior runs and existing annotation templates. Recent evidence suggests that LLM-as-judge alignment can vary with the grading scale, with competitive results reported under 0--5 settings~\cite{li2026grading}. We ran a retrospective pilot on the same 450-sample calibration pool without new API calls: affine relabeling from 1--5 to 0--5 was numerically invariant after inverse mapping (test MAE 0.4462 in both settings), while a quantized 0--5 integer simulation showed higher calibrated error (0.4870). This indicates that endpoint relabeling alone is unlikely to change outcomes, but score granularity can. A full online rerun with identical prompts and data, differing only in rubric anchors, is still required for definitive claims.

\paragraph{Few-Shot Prompting.}
Our experiments and reported results represent only zero-shot prompting approaches. However, preliminary experiments with few-shot prompting (prepending 1--3 expert critiques as in-context exemplars) yielded counter-intuitive results. Performance decreased with more examples, possibly due to attention dilution or style over-fitting. Future work should explore interventions that explicitly scaffold L1$\rightarrow$L5 reasoning, such as retrieval-augmented exemplars with semantic relevance.

% =============================================================================
% End of Limitations Section
% =============================================================================

% =============================================================================
% References
% =============================================================================
% \bibliographystyle{acl_natbib}
\bibliography{references_acl2026}

% =============================================================================
% Appendix (optional)
% =============================================================================
\appendix

\section{Robustness Analysis Details}
\label{app:robustness}

\subsection{Tier I--Tier II ICC Analysis}

Table~\ref{tab:icc_full} reports agreement between Tier~I automated metrics and Tier~II judge scores over 4,405 evaluations, including 95\% confidence intervals.

\begin{table}[ht]
\centering
\resizebox{\columnwidth}{!}{%
\begin{tabular}{lcccccc}
\toprule
\textbf{Tier I Metric} & \textbf{ICC(2,1)} & \textbf{95\% CI} & \textbf{TI Mean} & \textbf{TII Mean} & \textbf{Gap} & \textbf{p-value} \\
\midrule
DCR (auto) & 0.020 & [-0.01, 0.05] & 4.12 & 3.27 & $+0.85$ & $<10^{-30}$ \\
CSA (auto) & 0.174 & [0.14, 0.21] & 2.89 & 3.96 & $-1.07$ & $<10^{-30}$ \\
CDS (auto) & 0.179 & [0.15, 0.21] & 3.02 & 3.91 & $-0.89$ & $<10^{-30}$ \\
LQS (auto) & $-0.168$ & [-0.20, -0.13] & 3.85 & 4.32 & $-0.47$ & $<10^{-30}$ \\
\bottomrule
\end{tabular}}
\caption{Tier I--Tier II ICC analysis. Tier I = automated metrics (4); Tier II = judge rubric mean (5 metrics). All ICC values indicate poor agreement ($<0.5$), consistent with distinct measurement targets. Gap = TI $-$ TII.}
\label{tab:icc_full}
\end{table}

\paragraph{ICC Model Type Justification.}
We use ICC(2,1) throughout (see \S\ref{sec:methodology} for model selection rationale and reporting conventions).

\subsection{Rating Scale Sensitivity (Retrospective Pilot)}
\label{app:scale_sensitivity}

Because recent work reports that LLM-judge alignment can vary with grading scales~\cite{li2026grading}, we ran a lightweight retrospective pilot on the same 450-sample judge-human pool (295 train / 155 test) using the existing Claude outputs, without additional API calls.

\begin{table}[ht]
\centering
\footnotesize
\resizebox{\columnwidth}{!}{%
\begin{tabular}{lcccccc}
\toprule
\textbf{Regime} & \textbf{$n$} & \textbf{MAE$_\text{raw}$} & \textbf{MAE$_\text{cal}$} & \textbf{$\Delta$MAE} & \textbf{Spearman raw/cal} & \textbf{Kendall raw/cal} \\
\midrule
Native 1--5 (continuous) & 155 & 0.4539 & 0.4462 & +0.0077 & 0.4500 / 0.4493 & 0.3148 / 0.3147 \\
Affine 0--5 (continuous) & 155 & 0.4539 & 0.4462 & +0.0077 & 0.4500 / 0.4495 & 0.3148 / 0.3147 \\
Simulated 0--5 (integer bins) & 155 & 0.5358 & 0.4870 & +0.0488 & 0.3112 / 0.3112 & 0.2541 / 0.2541 \\
\bottomrule
\end{tabular}}
\caption{Retrospective scale-sensitivity pilot from existing scores. ``Affine 0--5'' is a continuous relabeling; ``Simulated 0--5'' applies integer quantization stress before calibration. This does not replace a true online rerun with 0--5 judge prompting.}
\label{tab:scale_sensitivity}
\end{table}

\paragraph{Takeaway.}
Changing endpoints by affine recoding alone is effectively invariant under sigmoid calibration in this dataset, while coarser discrete bins degrade raw and calibrated fit. Therefore, decisive claims about 1--5 vs 0--5 require a controlled online rerun where prompt rubric anchors are actually changed.

\subsection{Calibration Baseline and Sample-Size Sensitivity}
\label{app:calibration_baseline}

To validate our sigmoid calibration choice and assess potential overfitting under limited data~\cite{niculescu2005predicting}, we compare our default sigmoid calibrator with an isotonic regression alternative on the same 450-sample pool (295 train / 155 test), without new API calls.

\begin{table}[ht]
\centering
\footnotesize
\begin{tabular}{lcc}
\toprule
\textbf{Method} & \textbf{MAE} & \textbf{Spearman $\rho$} \\
\midrule
Raw (no calibration) & 0.4539 & 0.4500 \\
Isotonic & 0.4493 & 0.4462 \\
Sigmoid (default) & \textbf{0.4462} & 0.4493 \\
\bottomrule
\end{tabular}
\caption{Held-out aggregate calibration comparison (n=155 test).}
\label{tab:calibration_baseline}
\end{table}

\begin{table}[ht]
\centering
\footnotesize
\resizebox{\columnwidth}{!}{%
\begin{tabular}{rccc}
\toprule
\textbf{$n_{\text{train}}$} & \textbf{Isotonic MAE} & \textbf{Sigmoid MAE} & \textbf{Gap (sig$-$iso)} \\
\midrule
50  & $0.4613 \pm 0.0234$ & $0.4566 \pm 0.0144$ & $-0.0048$ \\
100 & $0.4449 \pm 0.0142$ & $0.4497 \pm 0.0091$ & $+0.0048$ \\
200 & $0.4430 \pm 0.0101$ & $0.4472 \pm 0.0038$ & $+0.0043$ \\
295 & $0.4493 \pm 0.0000$ & $0.4462 \pm 0.0000$ & $-0.0030$ \\
\bottomrule
\end{tabular}}
\caption{Calibration learning curve on held-out MAE (30 repeats for $n \in \{50,100,200\}$; deterministic for $n=295$).}
\label{tab:calibration_learning_curve}
\end{table}

\paragraph{Takeaway.}
At full sample size ($n=295$), sigmoid calibration achieves lower held-out MAE than isotonic regression (0.446 vs.\ 0.449). At smaller training sizes ($n \leq 200$), isotonic occasionally outperforms sigmoid due to its greater flexibility, while sigmoid maintains lower variance across repeats. We adopt sigmoid calibration for the main leaderboard because of its parametric parsimony (2 parameters vs.\ $n$ step points), guaranteed monotonicity, and smoother generalisation.

\subsection{Model-Stratified Cultural Gap Distribution}
\label{app:model_gap_distribution}

Across the 15-model panel, the mean China--West gap is $\Delta_{\text{CN-WE}}=-0.390$ with bootstrap 95\% CI $[-0.578,-0.230]$ (1,000 resamples; paired over models), consistent with a persistent Western advantage under our current evaluation setup.

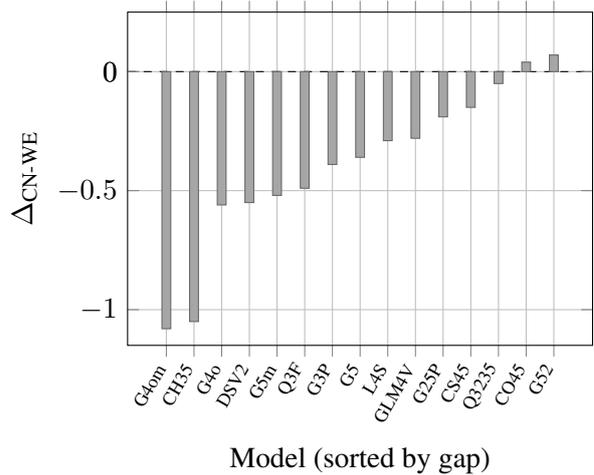
\begin{figure}[t]
\centering
\begin{tikzpicture}
\begin{axis}[
width=\columnwidth,
height=6.0cm,
ybar,
bar width=3.3pt,
ymin=-1.15,
ymax=0.25,
ylabel={$\Delta_{\text{CN-WE}}$},
xlabel={Model (sorted by gap)},
xtick=data,
xticklabel style={rotate=60,anchor=east,font=\scriptsize},
symbolic x coords={G4om,CH35,G4o,DSV2,G5m,Q3F,G3P,G5,L4S,GLM4V,G25P,CS45,Q3235,CO45,G52},
grid=major,
extra y ticks={0},
extra y tick style={grid style={dashed,black}},
]
\addplot[fill=black!35,draw=black!60] coordinates {
(G4om,-1.08)
(CH35,-1.05)
(G4o,-0.56)
(DSV2,-0.55)
(G5m,-0.52)
(Q3F,-0.49)
(G3P,-0.39)
(G5,-0.36)
(L4S,-0.29)
(GLM4V,-0.28)
(G25P,-0.19)
(CS45,-0.15)
(Q3235,-0.05)
(CO45,0.04)
(G52,0.07)
};
\end{axis}
\end{tikzpicture}
\caption{Model-stratified China--West gap distribution ($\Delta_{\text{CN-WE}}$). Negative values indicate higher Western scores. 13/15 models are below zero.}
\label{fig:model_gap_stratified}
\end{figure}

\subsection{Confound Control Analyses}
\label{app:confound_control}

\paragraph{Genre-controlled subset.}
Because Chinese samples are predominantly landscape paintings (54\% landscape) while Western samples show greater genre diversity, the observed CN--WE gap could partially reflect genre-related difficulty differences. To test this, we restrict comparison to landscape paintings only, matching the dominant genre across both cultures. Within the landscape subset ($n_{\text{CN}}{=}300$, $n_{\text{WE}}{=}405$ instance-level evaluations), the CN--WE gap persists and in fact strengthens: $\Delta=-0.40$, Cohen's $d=-0.93$ (95\% CI $[-1.09, -0.78]$, $p < 0.001$). This exceeds the overall effect ($d=-0.74$), confirming that genre distribution does not confound the cultural gap.

\paragraph{Blind-culture evaluation.}
To test whether judge cultural bias inflates the Western advantage, we re-evaluate 50 critiques (25~CN + 25~WE from GPT-4o) under a \emph{blind} condition where the culture tag is removed from the judge prompt. Under standard conditions (culture tag present), the CN--WE gap on this subset is $\Delta_{\text{std}}=-0.54$ (Cohen's $d=-2.10$, 95\% CI $[-0.68, -0.40]$). Under blind conditions, the gap \emph{widens}: $\Delta_{\text{blind}}=-0.61$ (Cohen's $d=-2.32$, 95\% CI $[-0.75, -0.47]$). The $+$13\% gap increase under blind conditions suggests the standard judge slightly \emph{compensated} for cultural difficulty when the culture tag was visible, rather than favouring Western art. All five dimensions show consistent or wider gaps under blind conditions. (The large $|d|$ values reflect the compressed within-group variance of calibrated scores on a small pilot; the raw gap $\Delta$ is the more interpretable effect measure here.)

\section{Reproducibility Details}
\label{app:reproducibility}

\paragraph{Generation-side settings.}
Each VLM receives a fixed structured prompt with explicit fields \textit{title}, \textit{artist}, and \textit{culture}, together with mandatory L1--L5 coverage instructions. Culture tags are fixed to \{\texttt{chinese}, \texttt{western}, \texttt{japanese}, \texttt{korean}, \texttt{islamic}, \texttt{indian}\}. Default request settings across all models: temperature $= 0.7$, max\_tokens $= 4{,}096$, max\_retries $= 3$, exponential backoff base $= 2.0$\,s. Model-specific max\_tokens overrides are applied where the provider supports longer outputs (e.g., GPT-5: 8{,}000, GPT-5-mini: 4{,}000).

\paragraph{Judge-side settings.}
Tier-II scoring uses Claude-Opus-4.5 (\texttt{claude-opus-4-5-20251101}) with \texttt{max\_tokens} $= 800$ and a JSON-constrained 5-dimension rubric prompt (Coverage, Alignment, Depth, Accuracy, Quality; abbreviated \texttt{DCR/CSA/CDS/FAR/LQS}), comparing VLM critique against expert reference. Judge calls use Anthropic's default \texttt{temperature} $= 1.0$, with \texttt{top\_p} and \texttt{top\_k} left unset. Per Anthropic's documentation, even \texttt{temperature} $= 0$ does not guarantee deterministic outputs. The JSON-constrained integer rubric (five dimensions, 1--5 scale) substantially limits output variability.

\paragraph{Parse failures and retries.}
Judge output parsing is two-stage: first JSON extraction, then regex fallback. Missing dimensions default to 3.0. API exceptions also return neutral defaults (3.0 per dimension) and are flagged for retry. Database writes are idempotent, and resume logic only skips successful rows, enabling reruns to repair failed items.

\paragraph{Versioning and fingerprints.}
Paper-code snapshot: \texttt{b51071d}.\footnote{\url{\frameworkurl}} Dataset fingerprint: \texttt{b8a34e5f}.

\section{Human Calibration Annotation}
\label{app:annotation}

This appendix details the human calibration annotation used in Tier~III (\S\ref{sec:methodology}). Expert reference critiques are drawn from the \textsc{Vulca-Bench} corpus; see \citet{yu2026vulca} for the full dataset construction and annotation protocol.

\paragraph{Task and Materials.}
For the 450-sample calibration subset, three annotator pairs independently score VLM-generated critiques on the five Tier~II dimensions using the rubric in Table~\ref{tab:scoring_rubric}. Each annotator receives (i)~the artwork image, (ii)~the expert reference critique, and (iii)~the VLM-generated critique, then assigns scores on a 1--5 scale with 0.5-step increments per dimension.

\begin{table}[ht]
\centering
\footnotesize
\begin{tabular}{@{}lp{5.2cm}@{}}
\toprule
\textbf{Dimension} & \textbf{Scoring Criterion} \\
\midrule
DCR & Does the VLM critique cover the same analytical dimensions as the expert reference? \\
CSA & Does the critique use culturally appropriate terminology and align semantically with the expert? \\
CDS & Does the critique reach comparable analytical depth (L1--L5 layers) as the expert? \\
FAR & Are factual claims (artist, period, technique, symbolism) accurate? \\
LQS & Is the critique linguistically fluent, coherent, and well-structured? \\
\bottomrule
\end{tabular}
\caption{Tier~II human scoring rubric (1--5 scale, 0.5-step increments). Annotators compare VLM critique against expert reference.}
\label{tab:scoring_rubric}
\end{table}

\paragraph{Assignment Design.}
Annotator pairs were assigned via a balanced incomplete block design (\S\ref{sec:methodology}): each pair scored a distinct subset of items independently on all five dimensions, yielding 299 dual-rated items for agreement analysis. This design ensures broad coverage of cultures and models while controlling for annotator fatigue.

\paragraph{Agreement Analysis.}
Overall weighted kappa was moderate ($\kappa_w = 0.43$, 95\% CI $[0.39, 0.45]$). Surface-oriented dimensions (Coverage, Quality) showed higher agreement ($\kappa_w = 0.54$ $[0.48, 0.58]$) than dimensions requiring cultural interpretation (Alignment, Depth, Accuracy: $\kappa_w = 0.35$ $[0.31, 0.39]$), consistent with the inherent subjectivity of L3--L5 assessment.

\section{5-Level Framework Dimensions}
\label{app:dimensions}

The 165 culture-specific dimensions were developed through an iterative process grounded in art-historical and cultural-theory literature. For each tradition, we conducted a literature survey of established analytical frameworks (e.g., Panofsky's iconological method for Western art, Six Principles of Chinese painting theory), then performed pilot tagging on a sample of expert critiques to identify recurring analytical categories. The resulting codebook was refined through two revision rounds and reviewed by domain-trained annotators. Full construction details, including inter-annotator agreement on dimension applicability, are reported in \citet{yu2026vulca}. Below we list each dimension with representative keywords; a dimension is considered \textit{covered} when the critique addresses its analytical content (e.g., CN\_L5\_D1 ``artistic conception'' requires discussion of mood or atmosphere beyond literal description, not merely mentioning the term \begin{CJK}{UTF8}{gbsn}意境\end{CJK}).

Table~\ref{tab:dims_overview} summarizes the 165 culture-specific dimensions across six traditions. Each culture has L1--L5 levels.

\begin{table}[ht]
\centering
\footnotesize
\begin{tabular}{@{}lcccccc@{}}
\toprule
\textbf{Culture} & \textbf{L1} & \textbf{L2} & \textbf{L3} & \textbf{L4} & \textbf{L5} & \textbf{Total} \\
\midrule
Chinese (CN) & 6 & 6 & 6 & 6 & 6 & 30 \\
Western (WE) & 6 & 6 & 5 & 4 & 4 & 25 \\
Japanese (JP) & 6 & 6 & 5 & 5 & 5 & 27 \\
Korean (KR) & 5 & 5 & 5 & 5 & 5 & 25 \\
Islamic (IS) & 5 & 5 & 6 & 6 & 6 & 28 \\
Indian (IN) & 6 & 6 & 6 & 6 & 6 & 30 \\
\midrule
\textbf{Total} & 34 & 34 & 33 & 32 & 32 & \textbf{165} \\
\bottomrule
\end{tabular}
\caption{Culture-specific dimension counts by level.}
\label{tab:dims_overview}
\end{table}

\paragraph{Chinese (CN, 30 dims).}
\textbf{L1} (Visual): colour usage, composition, brushstroke texture, spatial levels, line expression, negative space (\begin{CJK}{UTF8}{gbsn}留白\end{CJK}).
\textbf{L2} (Technical): \textit{cun} texture strokes (\begin{CJK}{UTF8}{gbsn}皴法\end{CJK}), brush techniques, ink gradation, colouring, brush dynamics, inscriptions/seals.
\textbf{L3} (Symbolic): imagery interpretation, symbolic meaning, cultural motifs, poetry-painting unity, literati aesthetics, aesthetic ideals.
\textbf{L4} (Historical): dynastic style, artist background, lineage, school characteristics, innovation, legacy.
\textbf{L5} (Philosophical): \textit{yijing} (\begin{CJK}{UTF8}{gbsn}意境\end{CJK}, artistic conception), \textit{qiyun shengdong} (\begin{CJK}{UTF8}{gbsn}气韵生动\end{CJK}, spirit resonance), heaven-human unity, aesthetic taste, form-spirit relation, void-solid interplay.

\paragraph{Western (WE, 25 dims).}
\textbf{L1}: colour palette, line quality, composition, perspective, texture, chiaroscuro.
\textbf{L2}: brushwork, colouring technique, medium/materials, subject/genre, style/school, technical mastery.
\textbf{L3}: iconography, symbolism, signature/inscription, patronage/function, literary references.
\textbf{L4}: period evolution, artist biography, influence/legacy, provenance.
\textbf{L5}: aesthetic ideals, sublime/emotion, cross-cultural influence, artistic value.

\paragraph{Japanese (JP, 27 dims).}
\textbf{L1}: colour, line, composition, spatial treatment, texture, light/shadow.
\textbf{L2}: brushwork, colouring, subject matter, school/style, period style, innovation.
\textbf{L3}: iconographic symbols, aesthetic concepts (\textit{wabi-sabi}, \textit{mono no aware}), inscriptions/seals, mounting format, aesthetic ideals.
\textbf{L4}: period characteristics, school evolution, artistic lineage, social context, artistic exchange.
\textbf{L5}: \textit{wabi-sabi} (imperfect beauty), \textit{mono no aware} (pathos of things), \textit{y\=ugen} (profound grace), narrative aesthetics, artistic philosophy.

\paragraph{Korean (KR, 25 dims).}
\textbf{L1}: colour usage, line expression, composition, spatial treatment, spirit resonance (\begin{CJK}{UTF8}{mj}기운\end{CJK}).
\textbf{L2}: brush techniques, ink methods, texture strokes, material properties, mounting format.
\textbf{L3}: landscape imagery, Four Gentlemen, Confucian-Daoist spirit, \textit{minhwa} folk painting, court painting.
\textbf{L4}: dynastic periods (Goryeo/Joseon), school lineage, artists, Sino-Korean exchange, provenance.
\textbf{L5}: literati aesthetics, \textit{jingyeong} (true-view landscape), genre aesthetics, Korean identity, artistic value.

\paragraph{Islamic (IS, 28 dims).}
\textbf{L1}: colour (lapis lazuli blue, vermilion), line quality, composition, spatial treatment, decorative patterns (arabesques).
\textbf{L2}: \textit{neg\=arg\=ari} technique, pigment preparation, calligraphy (\textit{nasta`l\=iq}), paper craft, preservation.
\textbf{L3}: epic narrative (\textit{Sh\=ahn\=ameh}), court life, religious themes, garden imagery (\textit{Ch\=ar B\=agh}), architecture, natural symbols.
\textbf{L4}: Persian schools (Herat, Tabriz, Shiraz), Mughal fusion, Ottoman tradition, Central Asian, dynastic patronage, manuscript tradition (\textit{kit\=abkh\=aneh}).
\textbf{L5}: calligraphic aesthetics, decorative philosophy, narrative aesthetics, Sufi spirituality, paradise vision, light symbolism.

\paragraph{Indian (IN, 30 dims).}
\textbf{L1}: colour (mineral pigments, gold), line, composition, spatial, figure style, decorative elements.
\textbf{L2}: miniature technique, pigment craft, paper (\textit{wasli}), brushwork, preservation, signature.
\textbf{L3}: Hindu iconography (Krishna, Radha), \textit{R\=agam\=al\=a} (musical modes), romance narrative, court life, natural symbols, mythology.
\textbf{L4}: Mughal court, Rajasthani schools, Pahari style, Deccan tradition, folk art, Company period.
\textbf{L5}: \textit{Rasa} theory (nine aesthetic emotions), \textit{Darshan} tradition (sacred seeing), \textit{Bhakti} devotion, decorative aesthetics, narrative aesthetics, spiritual realm.

\section{Case Study: Success and Failure Patterns}
\label{app:case_study}

\begin{table*}[ht]
\centering
\small
\caption{VLM critique excerpts illustrating success vs.\ failure patterns. Score = calibrated $S_{\text{II}}^*$ (1--5; higher is better). Success cases demonstrate L3--L5 cultural depth with technique terminology and historical context, while failure cases remain at surface-level description despite fluent language.}
\label{tab:case_study}
\begin{tabular}{@{}m{2.2cm}m{1.8cm}cm{6.0cm}m{3.4cm}@{}}
\toprule
\textbf{Image} & \textbf{Artwork} & \textbf{Score} & \textbf{Critique Excerpt} & \textbf{Analysis} \\
\midrule
\includegraphics[width=2.0cm]{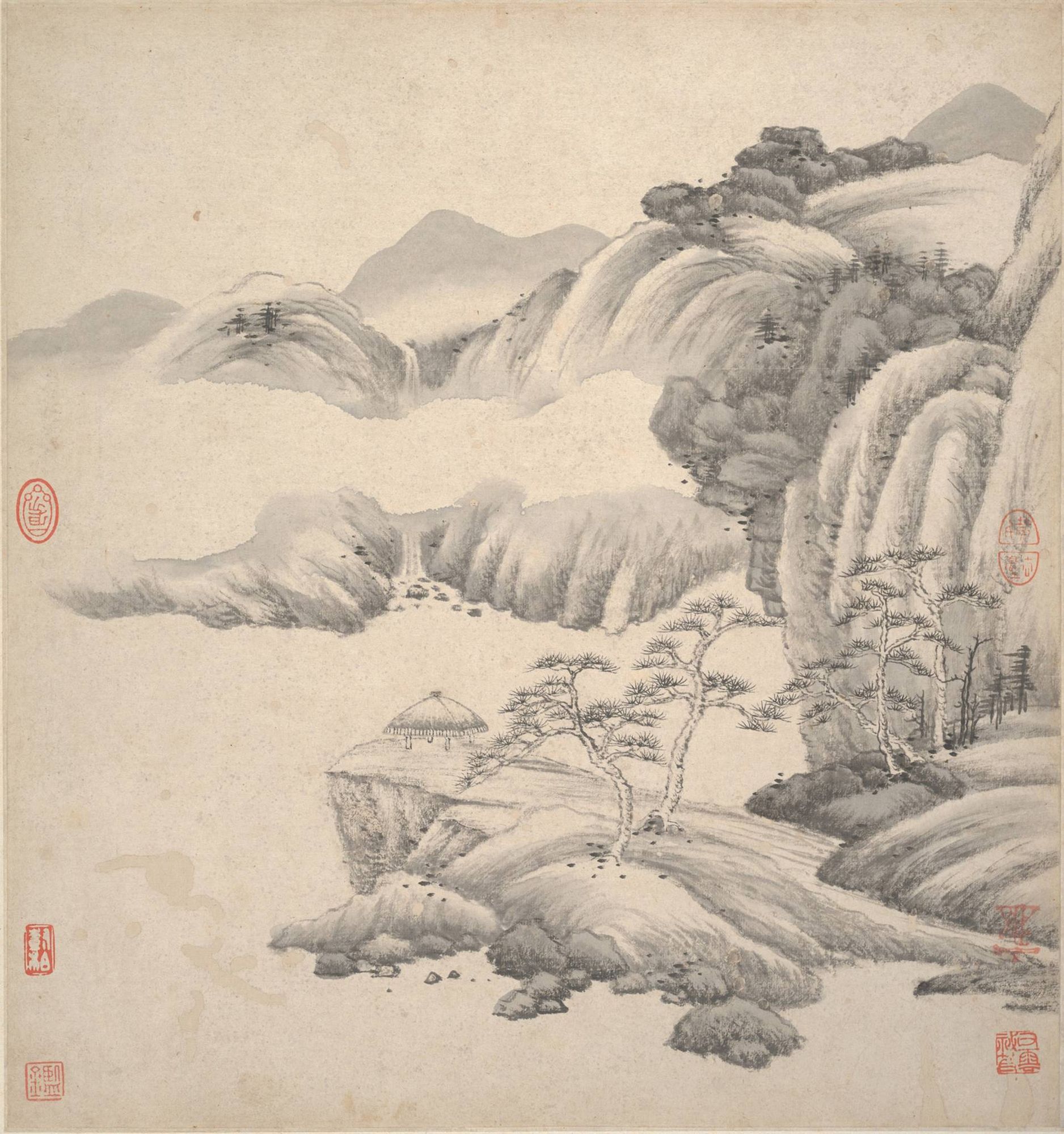} &
Wang Jian \newline \textit{Landscape Album} &
4.80 &
\textit{``...rocks rendered with \underline{pima cun} (hemp-fiber strokes), combined with \underline{short pima} and \underline{jiesuo cun}...embodying the \underline{Song-Yuan landscape tradition}''} &
\textcolor{green!50!black}{\textbf{Success}} \newline \textbf{L2}: technique terms \newline \textbf{L4}: historical lineage \\
\midrule
\includegraphics[width=2.0cm]{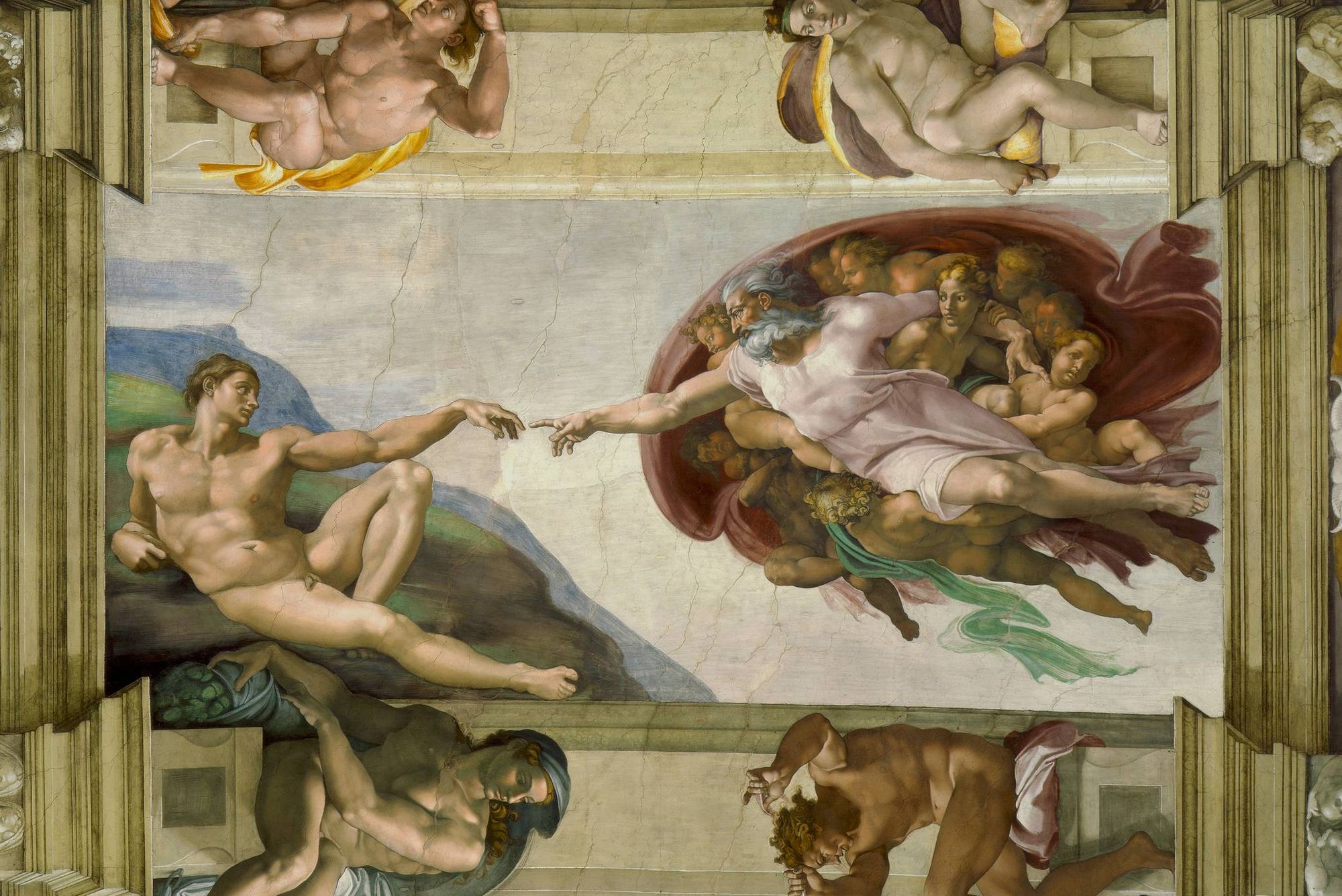} &
Michelangelo \newline \textit{Creation of Adam} &
4.95 &
\textit{``...this `\underline{near-touch yet not quite touching}' moment becomes the dramatic climax, \underline{symbolizing} life not yet fully awakened''} &
\textcolor{green!50!black}{\textbf{Success}} \newline \textbf{L3}: symbolic reading \newline \textbf{L5}: theological depth \\
\midrule
\includegraphics[width=2.0cm]{case3_shenzhou.jpg} &
Shen Zhou \newline \textit{Farewell at Jingjiang} &
2.45 &
\textit{``This work by Ming dynasty Shen Zhou uses ink technique...composition is reasonable, brushwork is exquisite...overall artistic conception is profound''} &
\textcolor{red!60!black}{\textbf{Failure}} \newline \textbf{Template}: generic praise \newline \ding{55} No specific technique \\
\midrule
\includegraphics[width=2.0cm]{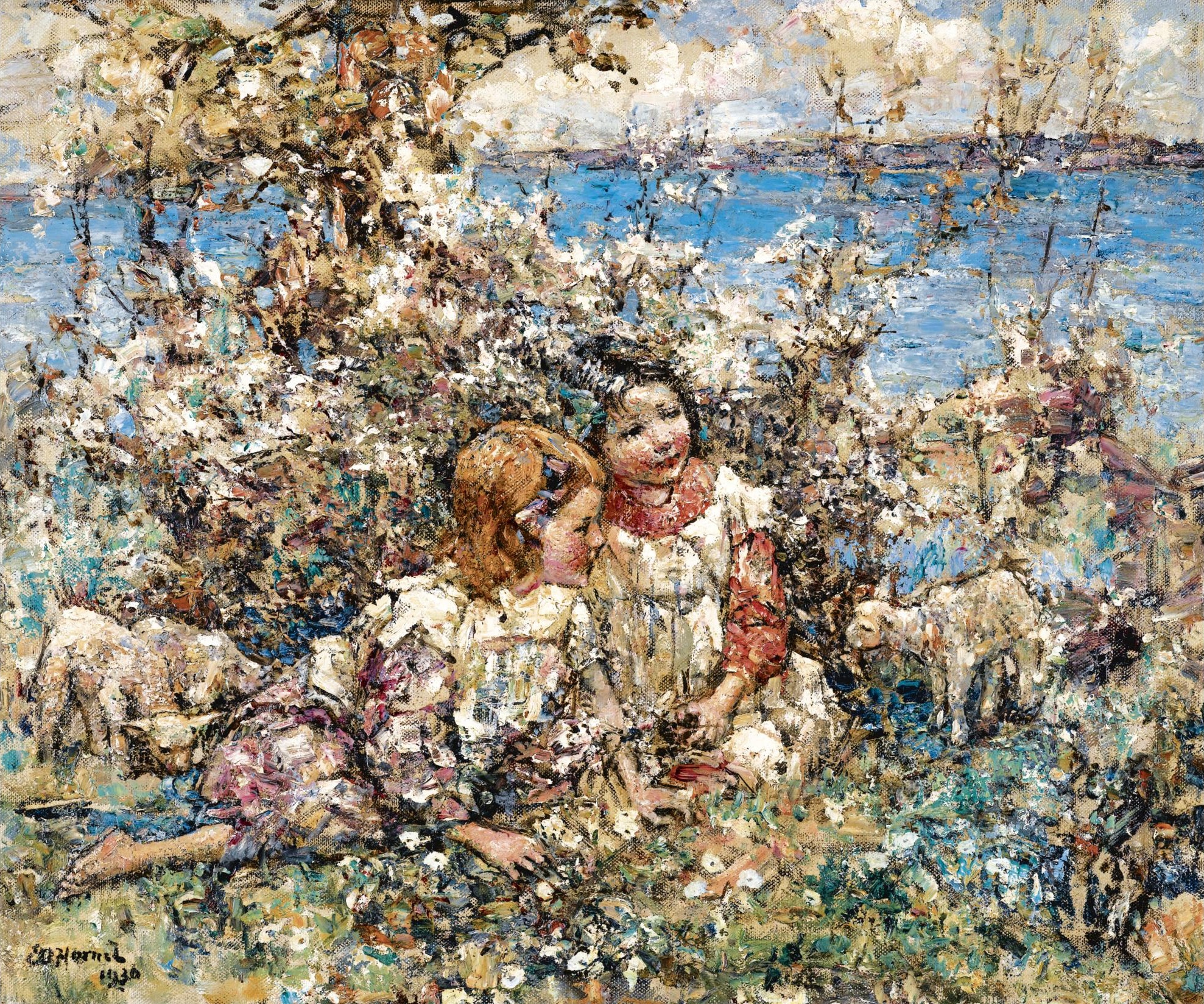} &
E.A.\ Hornel \newline \textit{Springtime} &
3.35 &
\textit{``Color relationships function on multiple levels...careful balance of forms establishes visual equilibrium''} &
\textcolor{red!60!black}{\textbf{Failure}} \newline \textbf{L1 only}: visual description \newline \ding{55} No Glasgow School context \\
\bottomrule
\end{tabular}
\end{table*}

\end{document}